\documentclass[authoryear]{elsarticle}
\usepackage{lineno}
\usepackage{hyperref}
\usepackage{lscape}
\usepackage{booktabs}
\usepackage{amsmath}
\usepackage{xr}
\usepackage{rotating}
\usepackage{multirow}
\usepackage[table]{xcolor}
\usepackage{graphics}
\usepackage{rotating}
\journal{Journal of Food Engineering}

\biboptions{authoryear}

\begin{document}

\begin{frontmatter}

\title{A new take on measuring relative nutritional density: The feasibility of using a deep neural network to assess commercially-prepared pur\' eed food concentrations}


\author[mymainaddress,secondaddress]{Kaylen J. Pfisterer \corref{mycorrespondingauthor}}
\cortext[mycorrespondingauthor]{Corresponding author}

\author[mymainaddress,secondaddress]{Robert Amelard}
\author[mymainaddress]{Audrey G. Chung}
\author[mymainaddress,secondaddress]{Alexander Wong}
\address[mymainaddress]{University of Waterloo, Department of Systems Design Engineering, N2L 3G1, Canada}
\address[secondaddress]{Schlegel-UW Research Institute for Aging, Waterloo, N2J 0E2, Canada}

\begin{abstract}
Dysphagia affects 590 million people worldwide and increases risk for malnutrition. Pur\' eed food may reduce choking, however preparation differences impact nutrient density making quality assurance necessary. This paper is the first study to investigate the feasibility of computational pur\' eed food nutritional density analysis using an imaging system. Motivated by a theoretical optical dilution model, a novel deep neural network (DNN) was evaluated using 390 samples from thirteen types of commercially prepared pur\' ees at five dilutions. The DNN predicted relative concentration of the pur\' ee sample (20\%, 40\%, 60\%, 80\%, 100\% initial concentration). Data were captured using same-side reflectance of multispectral imaging data at different polarizations at three exposures. Experimental results yielded an average top-1 prediction accuracy of 92.2\%$\pm$0.41\% with sensitivity and specificity of  83.0\%$\pm$15.0\% and 95.0\%$\pm$4.8\%, respectively. This DNN imaging system for nutrient density analysis of pur\' eed food shows promise as a novel tool for nutrient quality assurance.
\end{abstract}

\begin{keyword}
\texttt{nutrient sensing\sep deep learning \sep image processing \sep pur\' eed food}
\end{keyword}

\end{frontmatter}


\section{Introduction}
Dysphagia (swallowing difficulty) affects approximately 590 million people worldwide~(\cite{Cichero2016}) and at least 15\% of American older adults~(\cite{sura2012}) increasing these individuals' risk for malnutrition~(\cite{ilhamto2014,sura2012}). Malnutrition impacts quality of life~(\cite{keller2004}) and accounts for significant annual burden to the health care system of approximately \$15.5 billion in the United States~(\cite{goates2016}) and £7.3 billion in the UK~(\cite{russell2007}). Modified texture diets (e.g., pur\' eed food) have been used to allow safe ingestion of nutritional requirements in this population~(\cite{germain2006}). However, based on differences in preparation methods, nutrient composition can be highly variable ~(\cite{ilhamto2014}). This has practical implications especially for older adults with a generally lower intake; food must be as nutritious as possible to ensure adequate nutrient consumption. Additionally, pur\' ee thickness has safety implications; too thin a pur\' ee may cause choking~(\cite{ilhamto2014}). Thus, pur\' eed food quality assurance is required~(\cite{ilhamto2014}). 

There is currently a lack of tools to quantitatively and objectively assess the nutritional density of pur\' ees. To in part address this, international definitions for modified texture foods (including pur\' ee) were recently released by the International Dysphagia Diet Standardization Initiative (IDDSI)~(\cite{Cichero2016}).  However, implementation of these international definitions does not address nutrient density beyond pur\' ee consistency and adoption may be limited in practice. An automated imaging system may help reduce variance within or between human assessors due to differences in learning or experience; a seasoned pur\' ee cook has more intuition about what makes a safe and nutritious pur\' ee than a new cook~(\cite{ilhamto2014}). A system that can quantify the concentration of the pur\' ee could reduce cost and time while providing insight into nutrient density of a pur\' ee in health care settings.

Optical imaging systems provide a powerful solution to this problem. Specifically, these systems use the same type of information (visible optics); however, computational models provide objective and repeatable predictions. Borrowing from the field of biomedical optics, photon migration models have been used to estimate quantitative tissue properties such as blood oxygen saturation and hemoglobin concentration~(\cite{bigio2016}). Though primarily used in biomedical applications, these models provide a theoretical basis for quantitative nutritional assessment using optical imaging data. Additionally, recent advances in machine learning have been successfully applied to a vast range of fields from object recognition to pharmacy and genomics~(\cite{lecun2015}). Specifically, deep neural networks (DNNs) are biologically inspired by the visual cortex for decision making~(\cite{bengio2009}), and have been used with great success for specific complex tasks such as speech recognition~(\cite{hinton2012, dahl2012, hannun2014}), object recognition~(\cite{krizhevsky2012, he2015, lecun2004, simonyan2014}), and natural language processing~(\cite{bengio2003, collobert2008}). In image classification and other applications, however, there is often insufficient training data to properly train a conventional DNN due to nature of supervised learning which require a large number of network parameters and an abundance of labeled training data. In the case of pur\' eed food analysis, data insufficiency becomes a prominent concern due to the limited amount of available labeled data. Labeled data requires the acquisition of spectral and texture information of the pur\' eed food via imaging, and the cumbersome manual labeling process of the images by trained personnel. 

In this paper, we assess the feasibility of computational nutritional density analysis using an imaging system to provide feedback without the need for human assessor input. This preliminary dilutions study is motivated by the end-goal of nutrient density assessment. Using relative water concentration to initial concentration (i.e., pure commercially prepared product), we prepare a dilution series to observe the effect of relative increased water content on optical properties (color information, texture information, satruation etc.) for the purpose of determining the feasibility of using an optical imaging techniques for discrimination. Instead of traditional supervised learning, we use stacked autoencoders with a final softmax layer for dilution classification (i.e., discriminating between 20\%, 40\%, 60\%, 80\%, 100\% initial concentration). Autoencoders are DNNs that leverage unsupervised learning to provide a robust solution that is generalizable and extensible without compromising performance to complete a specific task. Specifically, this is the first study to our knowledge to assess the feasibility of using machine learning (DNNs) to automatically predict the concentration (as a proxy for nutrient density) of  commercially-prepared pur\' ees. Furthermore, the use of DNNs for this task is motivated by the results of a theoretical optical dilution model. In particular, since neural networks are biologically inspired machine learning methods and since in practice, food and food quality are often visually assessed, a theoretical optical validation of perceptually quantifiable nutrition composition can provide strong support for using machine learning. For example, passing input, such as a hypothetical concentration into a theoretical model, would yield an ideal output similar to the perception of the human eye. This present study, involving visible spectrum multispectral imaging data at different polarizations, provides a novel application of image classification to analyze thirteen types of commercially-prepared pur\' ees across three food categories (fruit, meat, vegetables) at five dilutions relative to initial concentration.

\section{Material and Methods}
\label{sec:Materials and Methods}
\setcounter{secnumdepth}{3}
\subsection{Sample preparation}
\label{ssec:Sample prep}
Thirteen commercially-prepared pur\' ee flavors across three food categories were selected for this study: fruit (apple, apricot, banana, blueberry, mango, strawberry), meat (beef, chicken), and vegetables (carrot, butternut squash, parsnip, pea, sweet potato). Pur\' ee flavors were selected to maximize variations in texture and color. For each pur\' ee, a five tier dilution series was prepared relative to initial concentration: 20\% (most diluted), 40\%, 60\%, 80\%, and 100\% (not diluted). For each dilution in the series, six 5 mL samples were systematically loaded onto a standardized transparency sheet grid from approximately one centimeter above the sheet at room temperature and imaged immediately, yielding a total of 390~samples.

\subsection{Data acquisition}
\label{ssec:Data acquisition}
Same-side reflectance was used (i.e., the light source and camera were positioned at the same location). A DSLR camera (Canon T4i) was used for high resolution image capture in the visible spectrum with consistent white balancing, aperture, and exposure settings. Both unpolarized and linearly polarized data were acquired by positioning an oriented linear polarizer in front of the camera lens. The use of polarization provided higher variability of a pur\' ee's appearance by focusing on surface-level texture (horizontal polarization) and color (vertical polarization) information. To simulate various lighting conditions, three exposures were acquired (1/20 s, 1/10 s, and 1/5 s) for each polarization. These variations enable the system to learn more robust concepts about the pur\' ees. Over the course of imaging, the room temperature varied from 21.9$^{\circ}$C to 23.9$^{\circ}$C.

\subsection{Sample subimages}
\label{ssec:Sample subimages}
Since neural networks are biologically inspired and food consistency is presently visually inspected, it may be helpful to describe the data in terms of tangible features such as color and texture. It is important to note that color and texture are meant only to provide intuition into the data collected and were not used as hand-crafted features; features used for distinguishing between classes (classification) were automatically learned given no priors through the deep neural network (see Section~\ref{ssec:Network arch} for more details). Figure~\ref{fig:sample_patches} provides a summary of color and texture across the samples. The images in Figure \ref{fig:sample_blobs} were acquired from the sixth sample location on the sheet. To minimize glare the horizontal polarization of entire sample subimages were selected to provide further context with an ISO 100 and exposure 1/20 s. 

\subsection{Training data set-up}
\label{ssec:Training data}
\begin{figure}
\centering
\includegraphics[width=\textwidth]{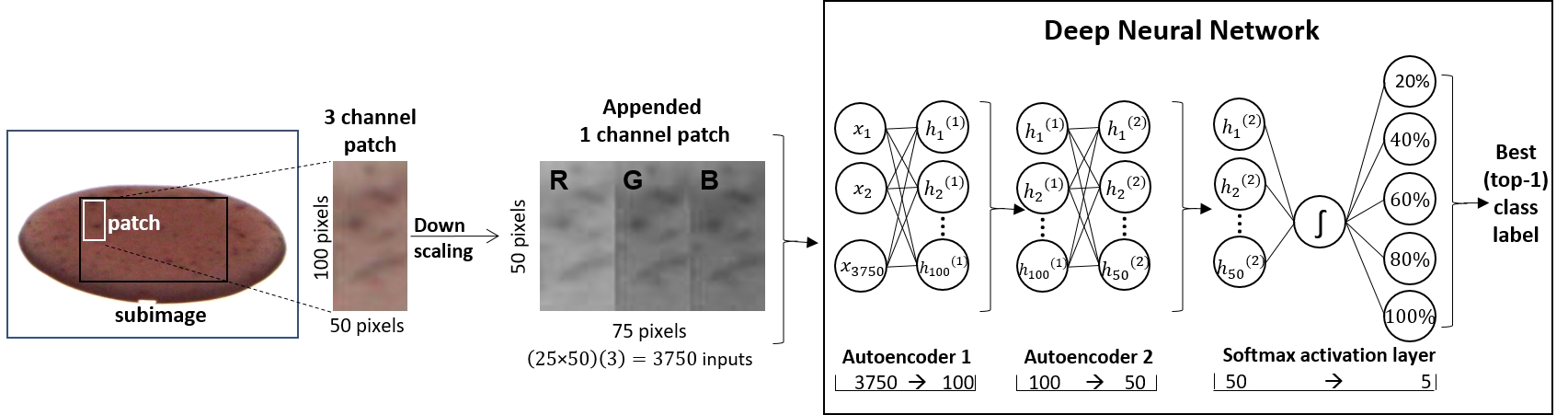}
\caption{Deep neural network architecture. A subimage is decomposed into a series of patches which are then downscaled to half their original dimensions (100$\times$50 --\textgreater 50$\times$25). RGB channels for each patch are concatenated to create one 50$\times$75 image per patch. Each patch is passed through two stacked autoencoders and a softmax layer to distinguish between classes (i.e., softmax classification layer). For a given flavor, the network output is one of five dilution classes (20\%, 40\%, 60\%, 80\%, and 100\%).}
\label{fig:network arch}
\end{figure}

Images were processed and data were analyzed using Mathworks’ MATLAB version R2016b. Each image was white normalized by selecting a reference white rectangle from an in-frame white reflectance target. All images were labeled and deconstructed into six, 100$\times$200 pixel subimages (one for each sample on the sheet). As indicated in Figure~\ref{fig:network arch}, each three channel (i.e., RGB) subimage was decomposed into fifty-four patches using half overlapping windows of 50$\times$100 pixels. Rectangular patches were selected to improve the variance observed within a patch. These patches were downscaled to 50\% of their original size (25$\times$50) using bicubic interpolation. The three RGB channels were concatenated to yield 378, 25$\times$50$\times$3 (or 75$\times$50) pixel patches for processing for a given dilution for a specific pur\' ee flavor and 1890 patches for a specific pur\' ee flavor. Therefore, the final set of images consisted of 13,230 auto-labeled patches.

\subsection{Network architecture}
\label{ssec:Network arch}
 
Images were then passed into a deep neural network (DNN), consisting of two layers of pretrained stacked autoencoders and a final softmax activation layer. At a high level, five global, general networks were formed using randomly initiated weights and passing through all of the unlabeled patches (i.e., there was no flavor or dilution information provided to the system). These general networks were then fine-tuned using flavor-specific labeled data. Given a specific flavor, the system predicted the dilution class to which a patch from an image belonged.

The autoencoders were implemented as feed-forward fully-connected implementations which use a logistic sigmoid function as a transfer function for nonlinearity. The input layer consisted of 3750 input neurons from image dimensions 50$\times$75. Similar to the process described by Hinton et al.~(\cite{hinton2006}) the first autoencoder layer was pretrained to receive the high-dimensional data from input layer and convert it to lower-dimensional data with 100 outputs via a learned nonlinear mapping. This large reduction in dimensionality was desirable based on the relatively basic characteristics of the data (e.g., texture and color). The output from the first autoencoder was passed into the second autoencoder in which dimensionality was further reduced to 50 outputs via a second learned nonlinear mapping, providing a ``distilled" feature set and back propagation was performed after training. Following the unsupervised learning from the stacked autoencoder, one final softmax layer was leveraged as a means to map the automatically learned ``distilled" features onto one of five output classes (20\%, 40\%, 60\%, 80\%, or 100\% initial concentration) as the top-1 hit (i.e., the one class with the strongest prediction) class label. Finally, the autoencoders and softmax layer were stacked together to form a general deep network where each autoencoder operates as a discrete feature extractor~(\cite{hinton2006}). This forced the network to learn increasingly higher level features. Running a final iteration of backpropagation across the whole system (i.e., the fine-tuning phase), we make use of the retained features that differentiate between classes~(\cite{hinton2006}) (i.e., inherent features that represent a concept such as the blueberry-ness of a specific concentration). From this global, general network, an additional step of fine-tuning was deployed for each flavor separately. This final fine-tuning step is the only iteration which uses the labeled, flavor-specific data. As a result, no hand-crafted features were used for the purpose of distinguishing between classes; all features were automatically discovered through the use of the stacked autoencoders.

\subsection{Validating our network: Pretraining and testing the network}
\label{ssec:Validating}
For each flavor-specific network, $k$-fold cross-validation was used by reserving one of the six positions for testing and completing training with the remaining five positions and conducting one final iteration of back propagation across the entire system for further fine-tuning of the weights. This was repeated five times for each flavor-specific network; specificity and sensitivity measures were averaged across each of the left-out positions. Accuracy of the network was assessed using confusion matrices whereby labels assigned by the network (i.e., observed class) were compared to the ground truth labels (i.e., expected class) and summarized by sensitivity and specificity for each class and across all classes for a given pur\' ee flavor.

\subsection{Comparative analyses}
\label{ssec:Comparative}
For comparative purposes, two methods were applied for feature extraction: 1) automatic extraction and learning of features by the second autoencoder, and 2) evaluation of hand crafted features based on color (64 features) and texture (seven features) characteristics~(\cite{zhang2012}). For further formulations of features, refer to Section~\ref{ssec:Theory comparative lvh}.

Once features were extracted (either from each of five general networks second autoencoders or from hand crafted features), each feature set for every patch was passed into one of three methods for distinguishing between dilution classes and making a prediction about to which dilution class a patch belonged: softmax layer, random forest, support vector machine (both linear and radial basis function kernel). The main difference in implementing these methods is that the random forest and support vector machine approach require features to be provided, while the softmax approach relies on automatically learned and generated features. For comparative purposes we provided the random forest and support vector machine approaches with hand-crafted features as well as the auto-generated features output from autoencoder 2. For further explanation of these machine learning methods, refer to Section~\ref{ssec:Theory comparative methods}.
\subsection{Data analyses}
\label{ssec:Data analyses}
Descriptive analyses were summarized based on accuracy at predicting concentration for a given pur\' ee flavor from confusion matrices. Texture was summarized using entropy. Entropy is a rotation-invariant statistical measure of disorder, and thus was used to quantify texture variation similar to other food classification studies (\cite{bosch2011, zhu2011}). In particular, local neighborhood entropy was used to assess the variation (or heterogeneity) of discrete image patches. Then, the spatial local neighborhood entropy distribution was used to summarize the texture of the entire image. Specifically, given a grayscale image $I$, local texture was computed as a region-wise neighborhood entropy computation:
\begin{equation}
H_i = -\sum_{x_j \in N_i} p_{N_i}(x_j) \cdot \log_2 (p_{N_i}(x_j))
\end{equation}
where $H_i$ is the local entropy of the $i^{\text{th}}$ pixel neighborhood $N_i$, $p_{N_i}$ is the intensity histogram (i.e., sample intensity probability distribution) of pixel neighborhood $N_i$, and $x_j \in N_i$ is the intensity value of pixel $j$. According to this formulation, samples with smooth homogeneous texture contain little intensity variation over localized patches, resulting in a sparse $p_{N_i}$ and thus low entropy. Conversely, samples with inhomogeneous or heterogeneous rough texture contains highly varying intensity values, resulting in a dense $p_{N_i}$ and thus high entropy. We used $9 \times 9$ pixel neighborhoods, resulting in 81 pixel intensities to populate a distribution containing 256 bins.

Color was summarized using the mean and standard deviation of red, green and blue values. Finally, saturation was summarized as a value between 0 and 1 where 1 represented totally saturated (white); saturation served as an indicator whether we could expect the system to work. If entropy was low and saturation was high, the data would represent pure white and may not contain discernible features upon which to correctly classify an image.

\section{Theory}
\label{sec:Theory}
\subsection{Comparative analyses: hand crafted features}
\label{ssec:Theory comparative lvh}
The color features were constructed using a discrete quantized color histogram. Color histograms are relatively invariant to rotation and translation, and coarse color quantization encourages perceptual similarities through enlarged bin sizes. Environmental consistency (e.g., exposure time, white correction, illuminant spectrum, etc.) is important for such color comparisons. A controlled optical setup was used to fix the relevant optical parameters, and is discussed further in Section \ref{ssec:Data acquisition}. Specifically, 64~color features were extracted by quantizing each color channel into four bins, yielding $4 \times 4 \times 4 = 64$ features. Given 64 colors, the number of pixels within a patch pertaining to each of the 64 color bins was counted. Normalized histograms were used such that the value for each bin in the histogram represented the percent of pixels belonging to that color bin. Mathematically, given image $I(x,y)$, the quantized values at each coordinate $q(x,y)$ was computed as:
\begin{equation}
q(x,y) = \text{arg}\min_i \left\{ || I(x,y) - c_i ||_2 \right\}
\end{equation}
where $c_i \in C$ is the set of uniformly spaced bins spanning the color space. The final quantized feature set was computed as:
\begin{equation}
f_c(i) = \sum_{(x,y) \in D} \delta(q(x,y),i)
\end{equation}
where $D$ is the image domain, and $\delta$ is the Kronecker delta function:
\begin{equation}
\delta(i,j) = \begin{cases}
1, & \text{if } i=j \\
0, & \text{if } i \ne j
\end{cases}
\label{eq:kronecker}
\end{equation}

With regards to texture features, we used a set of texture descriptors based on differential translation histograms~(\cite{zhang2012}). First, all patches were converted from color to grayscale, yielding $I_g(x,y) \in R$. Then, the translational sum and difference maps were computed:
\begin{align}
s_\Delta(x,y) &= I_g(x,y) + T_\Delta I_g(x,y) \\
d_\Delta(x,y) &= I_g(x,y) - T_\Delta I_g(x,y)
\end{align}
where $T_\Delta$ is the translation operator by coordinates $\Delta$. Then, the normalized sum and difference translation histograms were computed as:
\begin{align}
h_{s,\Delta}(i) = \frac{1}{N} \sum_{(x,y) \in D} \delta(s_\Delta(x,y),i) \\
h_{d,\Delta}(i) = \frac{1}{N} \sum_{(x,y) \in D} \delta(d_\Delta(x,y),i)
\end{align}
where $D$ is the image domain, $N$ is the number of pixels, and $\delta(\cdot)$ is the Kronecker delta function from Eq. (\ref{eq:kronecker}). These histograms represent texture descriptors. For example, pixels in homogeneous regions which exhibit low texture will be approximately equal, resulting in $s_\Delta(x,y) \approx 2I_g(x,y)$ and $d_\Delta(x,y) \approx 0$ for all $(x,y)$ in the homogeneous region.

Using these fundamental computations, the following texture features were computed for each patch~(\cite{zhang2012,unser1986}):
\begin{align}
\text{Mean: } &\mu = \frac{1}{2}\sum_i i h_{s,\Delta}(i) \label{eq:mean} \\
\text{Contrast: } &c_n = \sum_j j^2 h_{d,\Delta}(j) \\
\text{Homogeneity: } &h_g = \sum_j \frac{1}{1+j^2} h_{d,\Delta}(j) \\
\text{Energy: } &e_n = \sum_i h_{s,\Delta}(i)^2 \sum_j h_{d,\Delta}(j)^2 \\
\text{Variance: } &\sigma^2 = \frac{1}{2} \left( \sum_i (i-2\mu)^2h_{s,\Delta}(i) + \sum_j j^2 h_{d,\Delta}(j) \right) \\
\text{Correlation: } &c_r = \frac{1}{2} \left( \sum_i (i-2\mu)^2 h_{s,\Delta}(i) - \sum_j j^2h_{d,\Delta}(j) \right) \\
\text{Entropy: } &h_n = -\sum_i h_{s,\Delta}(i) \log(h_{s,\Delta}(i)) - \sum_j h_{d,\Delta}(j) \log(h_{d,\Delta}(j)) \label{eq:entropy}
\end{align}

\subsection{Comparative analyses: methods for distinguishing between dilution classes}
\label{ssec:Theory comparative methods}

As mentioned previously, three methods for distinguishing between dilution classes were used: a softmax layer, and for comparative purposes, random forests (\cite{breiman2001}), and support vector machines (SVM)~(\cite{cortes1995}). Here we describe a softmax layer. A softmax layer used the output features from the second autoencoder with the dilution labels to output the top-1 hit (i.e., the one class with the strongest prediction) class label. This method was applied to the features output from autoencoder 2 for each of the five general networks for the autoencoder based features only (i.e., hand crafted features were not fed into the softmax layer). $k$-fold cross-validation was applied to each iteration and results were averaged across the five networks.

\subsection{Optical dilution model}
\label{ssec:Optical dilution model}

An optical dilution model was developed to motivate the use of deep neural networks for dilution classification. As a photon traverses through the pur\' ee sample, it undergoes a series of scattering and absorption events according to the constituent chromophores, resulting in the perceived color. As the pur\' ee becomes diluted, the relative concentration of water increases while the photon path length stays relatively constant, leading to decreased overall absorption and thus changes in perceived color. Mathematically, expressing this using the Beer-Lambert law of light attenuation produces:
\begin{equation}
A = \log \left( \frac{I_0}{I} \right) = \epsilon_{H_2O} \cdot c_{H_2O} \cdot l_{H_2O} + \epsilon_{p} \cdot c_{p} \cdot l_{p}
\end{equation}
where $A$ is absorbance, $I_0$ and $I$ are the incident and reflected illumination respectively, $\epsilon_{H_2O}$ and $\epsilon_p$ are the chromophore extinction coefficients for water and pur\' ee, $c$ is the chromophore concentration, and $l$ is the mean photon path length through the absorbing medium. Assuming a homogeneous dilution mixture ($l_{H_2O} \approx l_{p}$), normalized incident illumination ($I_0=1$), and normalized relative concentration ($c_{H_2O}+c_{p}=1$), this formulation simplifies to:
\begin{equation}
A = -(\epsilon_{H_2O}(1-c_{p}) + \epsilon_{p}c_{p})l
\label{eq:A}
\end{equation}
where $l=l_{p}=l_{H_2O}$. Representative perceptual image patches were derived from the mixture absorbance spectra by computing the perceived spectra color according to the CIE LMS cone responsivity curves:
\begin{equation}
I_v = \int_{400}^{700} Z_v(\lambda) \cdot \exp\left( -(\epsilon_{H_2O}^{(\lambda)}(1-c_{p}^{(\lambda)}) + \epsilon_{p}^{(\lambda)}c_{p}^{(\lambda)})l \right) ~ d\lambda
\label{eq:I}
\end{equation}
where $I_v$ and $Z_v$ are the image and spectral cone response for color channel $v \in \{R,G,B\}$.

\section{Results}
\label{sec:Results}
To understand the performance of the deep neural network (DNN) for predicting pur\' ee sample concentration, results are organized as follows: (1) supporting evidence from the optical dilution model that dilution is quantifiable through perceptual data; (2) descriptive analyses of each image class in terms of color, texture and saturation to provide quantitative insights; (3) sample patches for each class across every pur\' ee flavor as a means to visualize and understand the underlying data; (4) an amalgamation of observations taken from confusion matrices to support accuracy of the system.

\subsection{Optical dilution model}
\label{ssec:Results optical}

\begin{figure}
\centering
\includegraphics[width=\textwidth]{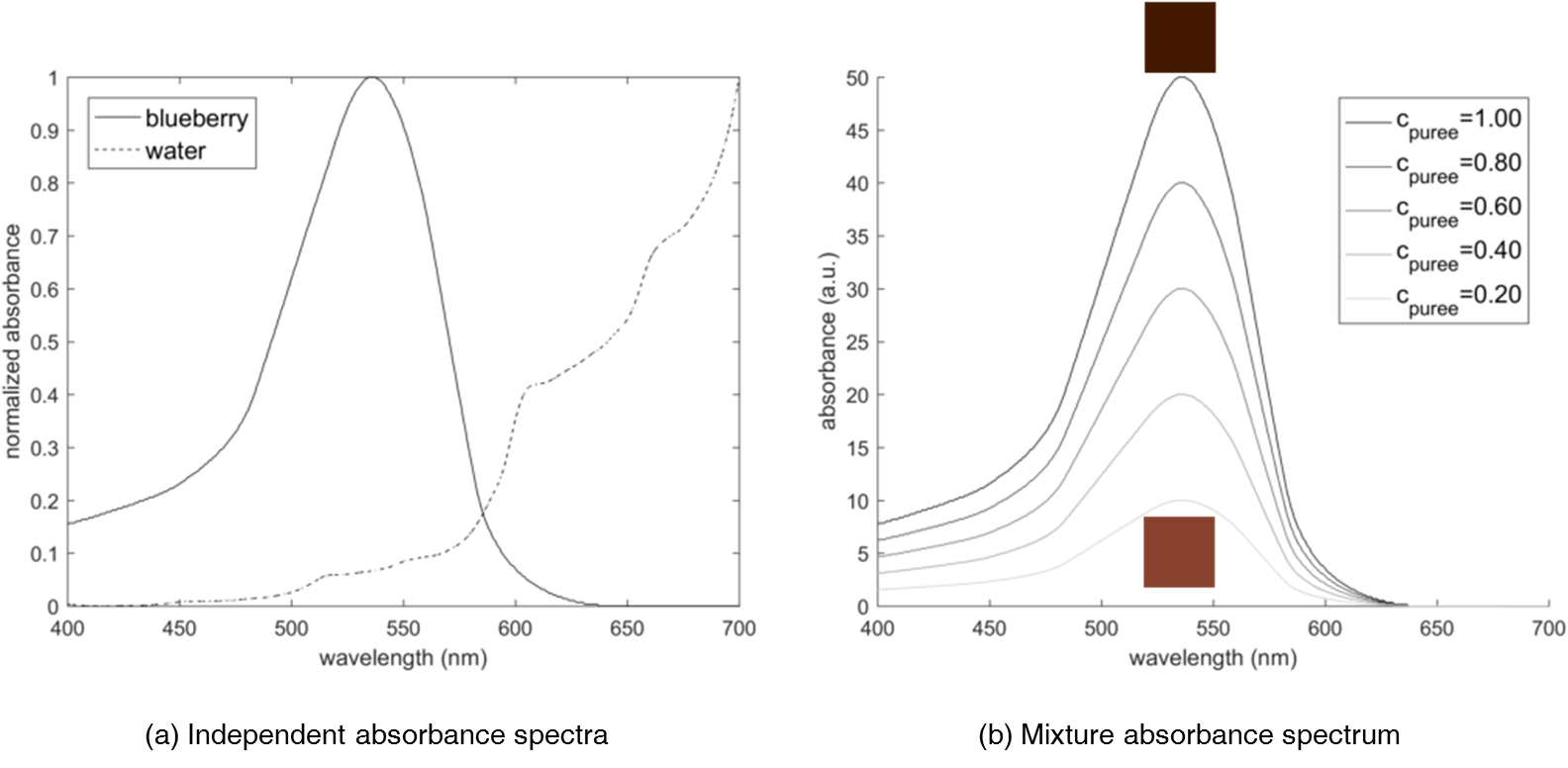}
\caption{(a) Normalized absorbance spectra for blueberry~(\cite{teoli2016}) and water~(\cite{robin1997}). (b) Effect of increasing dilution (decreasing relative pur\' ee concentration) on the water-blueberry mixture spectral curve. As pur\' ee dilution increases, the overall absorption decreases due to fewer photon absorption events during the photon migration path, leading to lighter observed images (bottom patch). These findings are valid for other absorbing pur\' ee chromophores without loss of generality.}
\label{fig:optics_change}
\end{figure}

The optical dilution model was evaluated with a candidate pur\' ee, blueberry, to motivate the use of neural networks as pur\' ee concentration estimators. Figure~\ref{fig:optics_change} demonstrates how the absorbance spectrum changes according to the pur\' ee concentration (i.e., dilution) according to (\ref{eq:A}) using published absorbance curves for water~(\cite{robin1997}) and blueberry~(\cite{teoli2016}). Blueberries contain anthocyanins which are pH sensitive chromophores which shift from red to blue with increasing pH. For example pH = 1 appear red, while at a pH = 4.5 they appear colorless and at pH 7-8 they appear blue ~(\cite{wrolstad1993}). As the undiluted blueberry pur\ ee becomes more diluted, its pH increases causing a shift from red to blue. In its undiluted state the spectral curve matches that of blueberry. As the pur\' ee becomes diluted with water, which has weak visible absorption, the mixture's absorbance decreases, resulting in an observable difference in spectral composition. These phenomena are reflected in the generated theoretical image patches in Figure~\ref{fig:optics_change} b according to (\ref{eq:I}). These findings support the hypothesis that pur\' ee  concentration can be quantifiably estimated using a perceptual machine learning framework; there is consistency between what is visually observed and what can be quantifiably described by the optical dilution model. Thus, DNNs, which leverage visually observable information and model complex non-linear relationships, seem to be a good model for predicting pur\' ee concentrations since they are biologically inspired and modeled after the human visual cortex for decision making~(\cite{bengio2009}).

\subsection{Descriptive analyses}
\label{ssec:descriptives}

Figures \ref{fig:RGB_norm} and \ref{fig:sat_entropy} provide qualitative information about each class of images with respect to color (mean R, G, B), texture (based on entropy, a statistical measure of variation) and saturation (where 0 is black and 1 is white). In terms of color, with the exceptions of banana and chicken the colors appeared more vibrant as percent initial concentration increased. This is intuitive since the lower percent initial concentration (i.e., more diluted) samples contained more water than their higher initial concentration counterparts, resulting in texture and surface tension more similar to water than the pure pur\' ee. While the samples were all imaged using the same lighting conditions, camera settings, and were white corrected, there was a large range of saturation, texture (entropy), and RGB values. The samples most at risk for oversaturation were the 20\% of initial concentration (IC) and the least at risk for oversaturation were the 100\% IC. At both the 20\% and 100\% dilutions, the most saturated samples were chicken (saturation: 20\% IC 0.992$\pm$0.008, 100\% IC 0.988$\pm$0.015) and the least saturated samples were blueberry (saturation: 20\% IC 0.538$\pm$0.071, 100\% IC 0.128$\pm$0.035). With respect to texture (entropy), note the specks of blueberry seeds in blueberry, the smooth shininess of banana, beef and chicken, the more granular surface texture in the butternut squash, and the consistent, fine granularity across the sweet potato classes as shown in Figure~\ref{fig:sample_patches}. In terms of texture (entropy) the more diluted samples were more similar in appearance to water and aside from their color, looked similar. Samples of lower dilution (more highly concentrated) tended to have higher entropy, however the most cohesive of samples (e.g., banana, beef, chicken, sweet potato) exhibited extremely smooth surface textures (i.e., lower entropy) across classes. This observation can be explained given that starches and proteins have a tendency to form gels (\cite{alvarez2013}) as these were the products with the highest starch contents (sweet potato: 5 g/128 mL, banana: 3 g/128 mL) or protein contents (beef: 12 g/10 0mL, chicken: 16 g/100 mL). For a full list of saturation and RGB values refer to Table~ \ref{tab:RGB_sat_ent}; for descriptive statistics pertaining to computational texture features (e.g., mean, contrast, homogeneity, energy, variance, correlation, and entropy), refer to Table~ \ref{tab:texture_features}.

\subsection{Sample patches}
\label{ssec:sample patches}

\begin{figure}
\centering
\includegraphics[width=\textwidth]{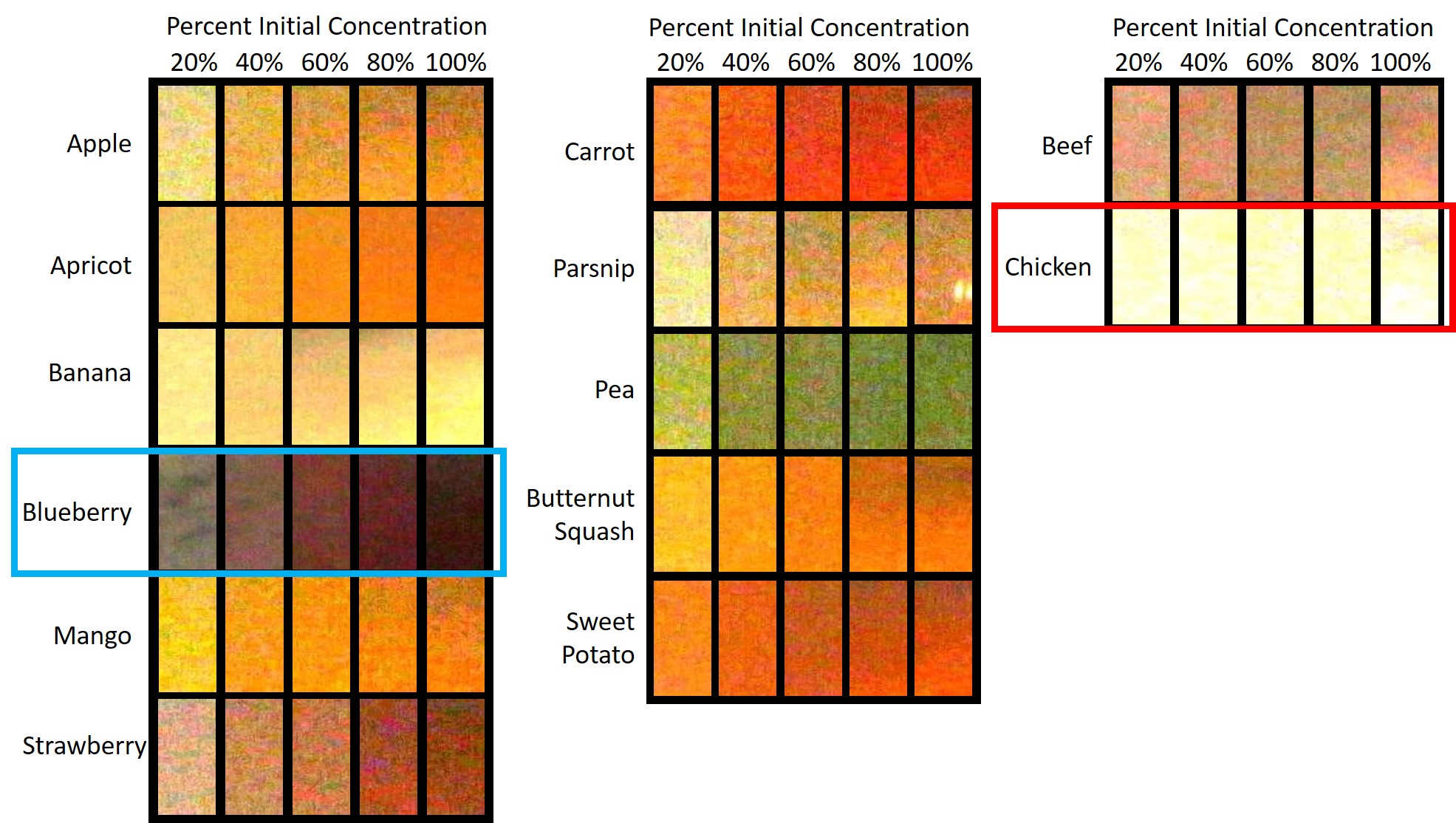}
\caption{Sample patches for each pur\' ee flavor and dilution. Note the visible color and texture variation across dilution classes and the indistinguishable nature of the poorest performing pur\' ee flavor highlighted in red. The sample blueberry patches outlined in blue had the best accuracy (99.6\%$\pm$0.6\%) with the autoencoders and softmax approach; these patches match the theoretical patches generated with the optical dilution model well (see Figure~\ref{fig:optics_change}).}
\label{fig:sample_patches}
\end{figure}

Figure \ref{fig:sample_patches} depicts sample patches for each class of pur\' ee flavor taken from the eighth patch generated from the first subimage. The sample blueberry patches match the theoretical patches generated using the optical dilution model in Figure~\ref{fig:optics_change} which supports the hypothesis that quantifiable observational evidence can be used to estimate relative nutrient concentration. A color intensity gradient across the concentrations was observed. Several pur\' ee flavors, most prominently in banana and beef, also exhibited a gradient across an image class most notably in the higher concentrations. For example from bottom to top, the 100\% beef samples darken. This was due to the highly cohesive nature of these samples; much more of the 5 mL sample loaded onto the sheet vertically rather than spreading horizontally. Specifically, this was due to the properties of the initial viscosity of the samples (i.e., there was variance in the viscosity of the 100\% initial concentration) not as a result of the preparation of the dilution series. These gradients are visual indications which the network may be using to distinguish between different concentration classes.

\subsection{Network accuracy}
\label{ssec:network accuracty}

The observations noted provide both quantitative (Figures \ref{fig:RGB_norm} and \ref{fig:sat_entropy}) and qualitative (Figure~\ref{fig:sample_blobs}) insight into performance. The method with the highest performance across flavors was our proposed DNN with an overall accuracy of 92.2\%$\pm$4.1\%, sensitivity of 83.0\%$\pm$1.5\%, and specificity of 95.0\%$\pm$4.8\%. This was closely followed by the handcrafted features paired with random forests for discrimination between dilutions. As illustrated in Tables~\ref{tab:AutoFeat},\ref{tab:HandFeat}, the most consistently highest performing pur\' ee flavor was strawberry. Pertaining to the most highly performing method for discrimination was the stacked autoencoders and softmax layer approach, blueberry performed best. Across 10 trials, the mean accuracy for classifying blueberry dilutions was 99.6\%$\pm$0.6\% (sensitivity 98.9\%$\pm$1.9\%, specificity 99.7\%$\pm$0.5\%). These results are consistent with the descriptive analyses based on the high variance  of color, entropy (texture) observed between classes of blueberry dilutions in addition to less image saturation across dilution classes. For example the lowest concentrations appeared more grey-blue compared to a more red-purple of the high concentration sample. Additionally, the blueberry samples also contained flecks of blueberry seeds or peels more visible in the lower concentrations than higher concentrations. These intuitive observations are congruent with the optical dilution model and the quantitative descriptive analyses with consistent and high accuracy. All other pur\' ee flavors' average accuracy ranged between 73.3\%$\pm$7.8\% (chicken) and 98.2\%$\pm$1.2\% (strawberry). Across all seven methods for discrimination between dilutions, chicken was the most difficult flavor which was reflected in the single poorest accuracy, sensitivity and specificity for every method. This was unsurprising since chicken samples were relatively indistinguishable to the human eye for the first several concentrations as they all simply looked white with no discernible features. This was consistent with the low entropy and high saturation seen in Table~\ref{tab:Summary}.

\begin{table}[htbp]
  \centering
  \caption{Summary of sensitivity, specificity and accuracy across all flavors using either self-generated features extracted from an autoencoder or color and texture based handcrafted features using four methods to discriminate between dilutions: softmax layer, random forest, SVM - linear kernel SVM, and SVM - radial basis kernel.}
	\begin{tabular}{lccc}
    \toprule
    \multicolumn{4}{c}{\textbf{SUMMARY OF PERFORMANCE ACROSS 13 FLAVORS}} \\ 
    \midrule
	\textbf{Method} & \textbf{Sens} & \textbf{Spec} & \textbf{Acc} \\
	\textbf{(AutoFeat)} & \textbf{($\mu_{Sens} \pm \sigma_{Sens}$)} & \textbf{($\mu_{Spec} \pm \sigma_{Spec}$)} & \textbf{($\mu_{Acc} \pm \sigma_{Acc}$)} \\
    \midrule
	SVM Linear & 0.690 $\pm$ 0.187 & 0.900 $\pm$ 0.074 & 0.844 $\pm$ 0.061 \\
    SVM Radial Basis & 0.609 $\pm$ 0.179 & 0.868 $\pm$ 0.082 & 0.794 $\pm$ 0.070 \\
    Random Forest & 0.790 $\pm$ 0.143 & 0.937 $\pm$ 0.046 & 0.903 $\pm$ 0.039 \\
    \textbf{Softmax} & \textbf{0.830 $\pm$ 0.150} & \textbf{0.950 $\pm$ 0.048} & \textbf{0.922 $\pm$ 0.041} \\
    \midrule
	\textbf{Method} & \textbf{Sens} & \textbf{Spec} & \textbf{Acc} \\
	\textbf{(HandFeat)} & \textbf{($\mu_{Sens} \pm \sigma_{Sens}$)} & \textbf{($\mu_{Spec} \pm \sigma_{Spec}$)} & \textbf{($\mu_{Acc} \pm \sigma_{Acc}$)} \\
    \midrule
	SVM Linear & 0.665 $\pm$ 0.188 & 0.890 $\pm$ 0.081 & 0.830 $\pm$ 0.057 \\
    SVM Radial Basis & 0.577 $\pm$ 0.175 & 0.850 $\pm$ 0.078 & 0.770 $\pm$ 0.063 \\
    Random Forest & 0.826 $\pm$ 0.158 & 0.949 $\pm$ 0.047 & 0.920 $\pm$ 0.044 \\
    \bottomrule
    \end{tabular}%
  \label{tab:Summary}%
\end{table}%

\section{Discussion}
\label{sec:Discussion}
While the highest performing method for discrimination between classes was the DNN, arguably, the combination of hand-crafted features with random forests for discrimination performed comparably. However, when generalizability is considered our DNN method using stacked autoencoders and a softmax layer provides several advantages. First, since autoencoders leverage unsupervised learning for the purpose of training, labels are unnecessary to create a global dilution model. Our initial global model contained all 13 flavors (flavors unlabeled) to learn the five dilutions. To fine tune this global model of dilution classes for a specific flavor a relatively small amount of labelled data for the new flavor would be required compared to forming a new flavor specific model as in the case of hand-crafted features with random forests for distinguishing between dilutions. Second, while it is unclear how either method would perform on more complex foods (e.g., regular texture, multiple food types on a plate), the deep learning approach may show more promise for extensibility since DNNs have historically be more flexible as evidenced by high accuracy across diverse applications (e.g., speech recognition~(\cite{hinton2012, dahl2012, hannun2014}), object recognition~(\cite{krizhevsky2012, he2015, lecun2004, simonyan2014}), and natural language processing~(\cite{bengio2003, collobert2008})). A third advantage is in the case of introducing a new flavor. Specifically, for the hand-crafted feature and random forest approach, no classification can be made without training a new model. Instead, our approach is based upon a general global dilution model and may therefore be used a starting point for discriminating between classes; initial accuracy could then be improved by fine-tuning the global dilution model for a new-flavor specific model. 

The subimages and patches used for this study reflected the 1/10 s exposure time. By re-running the same tests on the 1/20 s data, we may be able to improve performance. Since the color of the same sample compared between exposures was not equal it appears there is a fundamental difference with how the light interacts with the samples. It would be interesting to test this hypothesis by investigating whether there is a correlation between the degree of difference and composition and to explore whether there are correlations between color (and perhaps relative or normalized entropy) and composition of macronutrients (e.g., carbohydrates, protein, fat) or micronutrients (e.g., vitamin A, iron). Additionally, as part of future work, this computational nutritional density analysis should be validated with traditional rheology methods. For future extension, additional testing should be conducted using nutrient specific manipulations in which food components to determine whether the optical dilution model holds true for changes in substances extending beyond water content.

The system proposed here is the first step towards the end-goal of a ``smart'' imaging system to automatically detect the concentration and composition of commercially prepared pur\' eed samples. As such, the output from the system offered only the best label. Future work will explore the confidence of a particular class instead of simply the output (i.e., label) from the softmax layer. In addition, classification was performed on a patch by patch basis; it may be more meaningful to instead classify based on the entire subimage using either a weighted average with mean squared error as a measure of accuracy or even simply taking the mode class of patches. The end goal would be to provide a means for a cook, dietitian or caregiver to run an image through the pre-trained system and have a relative nutrient density estimate to inform nutritional density and texture safety measurements. Future directions of output from the system should be based on input from end-users to ensure output from the system is most clinically meaningful and relevant. In addition, since our system generalized well to new data, we expect this approach will allow for robust extensions including combining classification of pur\' ee flavor as well as dilution, and to extend beyond pur\' eed food to other modified textures (e.g., from regular to minced, to pur\' eed, and to liquidized) consistent with the recent development of the International Dysphagia Diet Standardization Initiative's dysphagia diet terminology~(\cite{Cichero2016}).

\section{Conclusions}
\label{sec:Conclusions}
In this paper, we demonstrated the feasibility of automatic nutritional density analysis using deep neural networks to predict the concentration of commercially prepared pur\' ees. Using multispectral imaging data at different polarizations, a stacked autoencoder attained promising accuracy across thirteen common pur\' eed foods. Of the 13 pur\' ee flavors tested, the most promising methodology was using DNNs comprised of stacked autoencoders with a softmax layer to distinguish between dilutions. Over all flavors and across 10 trials the mean accuracy of this method was 92.2\%$\pm$4.1\% with mean sensitivity 83.0\%$\pm$15.0\%, and mean specificity 95.0\%$\pm$4.8\%. Dilution classification results were strongest for pur\' ee flavors with observable texture differences reflected in higher entropy, higher variation in color across dilution classes, and lower saturation. In contrast, pur\' ee flavors performed more poorly when there were fewer visual cues to discriminate between dilution classes which was further reflected in low color variation, low entropy across classes, and high saturation. These findings begin to clarify the constraints of working towards classification with naturalistic images taken in the field. If accuracy can be further improved and the system can achieve similar accuracy on a wider range of modified texture foods, this machine learning DNN imaging system for nutrient density analysis of pur\' ees show promise as a tool for pur\' ee nutrient quality assurance.

\section*{Acknowledgements}
The authors would like to acknowledge Professor Heather Keller for her contribution to the overall vision of this developing system. This work was supported by the Natural Sciences and Engineering Research Council of Canada, the Canada Research Chairs Program, and Nvidia for the GPU hardware used in this study through the Nvidia Hardware Grant Program.

\section*{References}
\bibliography{mybibliography}

\newpage
\section*{Appendices}
\setcounter{equation}{0}
\renewcommand{\theequation}{A.\arabic{equation}}
\newcounter{atable}
\renewcommand{\thetable}{A.\arabic{atable}}
\newcounter{afigure}
\renewcommand{\thefigure}{A.\arabic{afigure}}

\refstepcounter{afigure} 
\begin{figure}
\centering
\includegraphics[width=\textwidth]{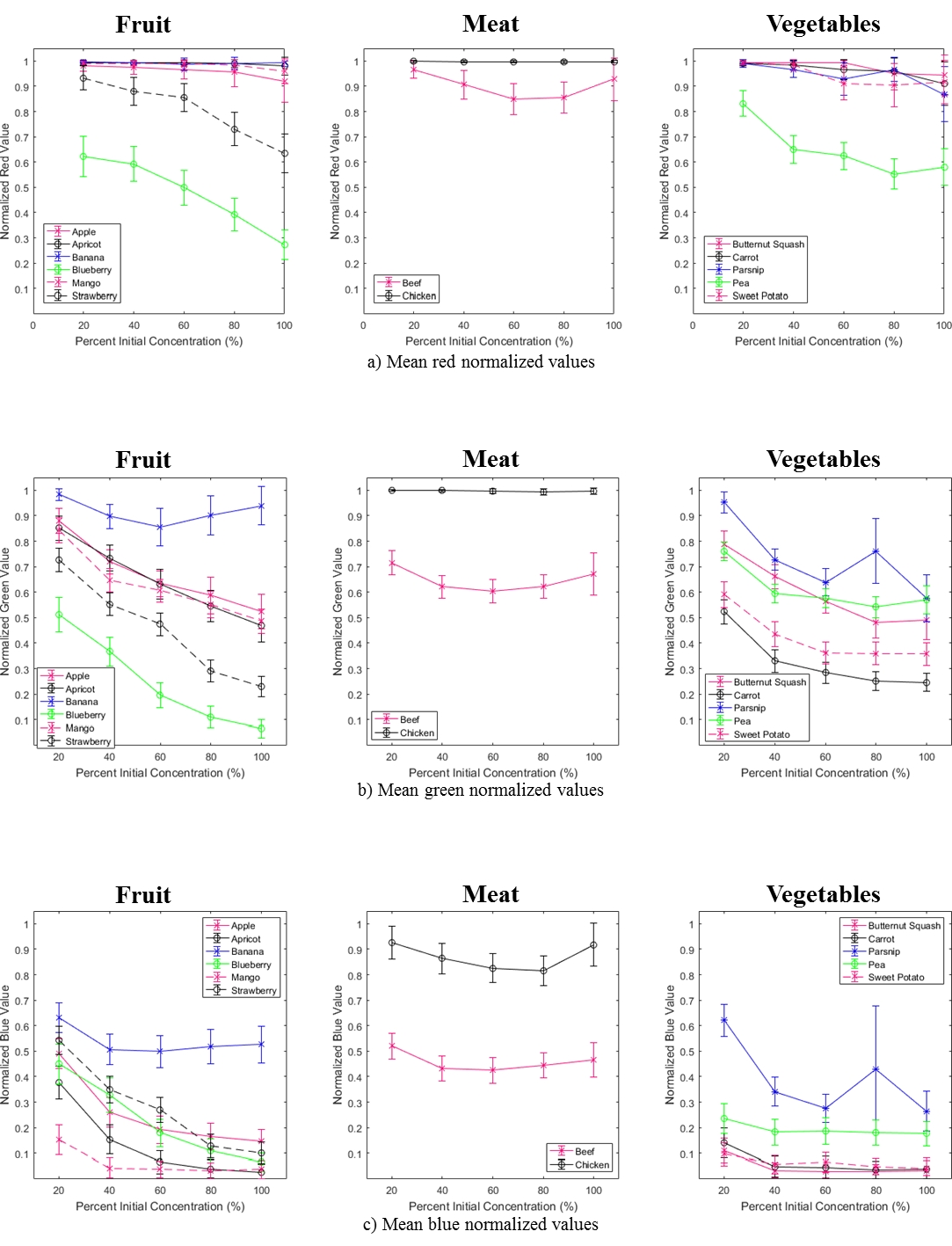}
\caption{Descriptive analysis plots of pur\' ee flavors based on color. RGB values  have been normalized. Typically, color varied between the dilution classes of a pur\' ee flavor and were distinguishable between different pur\' ee flavors.}
\label{fig:RGB_norm}
\end{figure}

\refstepcounter{afigure} 
\begin{figure}
\centering
\includegraphics[width=\textwidth]{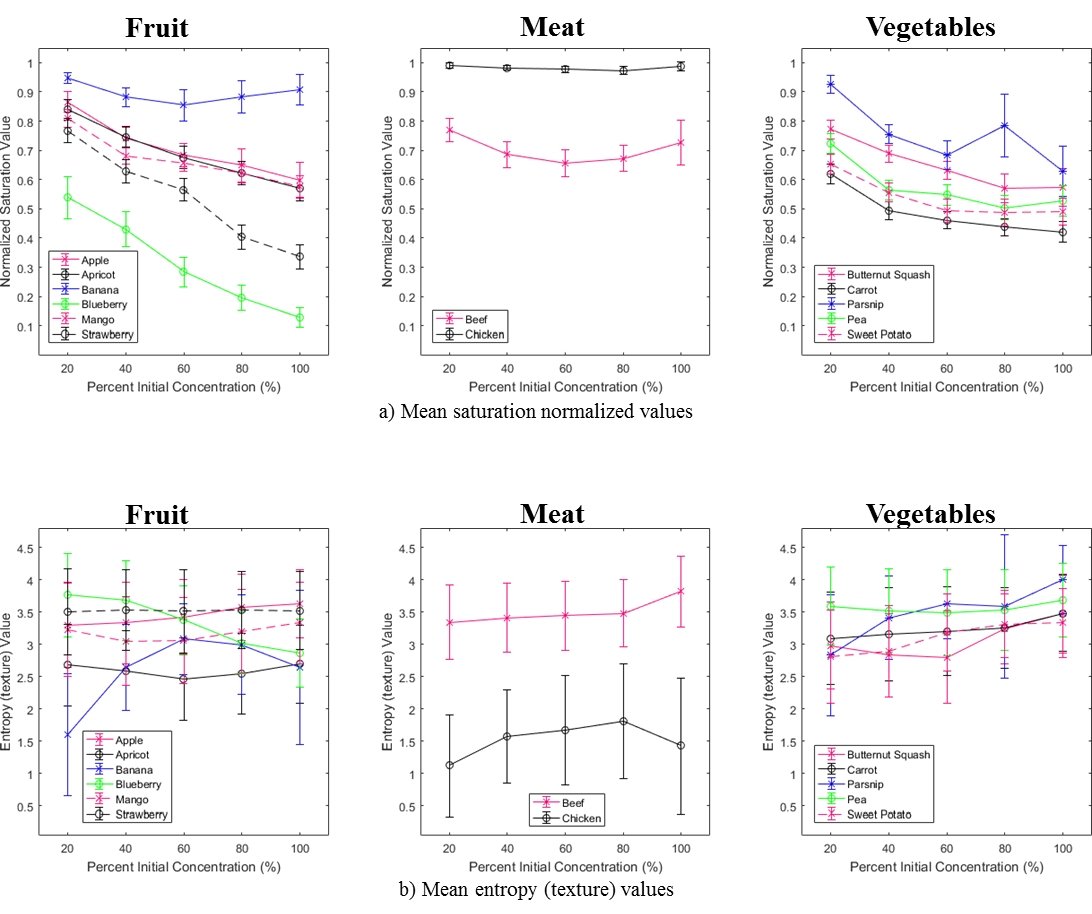}
\caption{Descriptive analysis plots of pur\' ee flavors based on saturation and texture (entropy). Saturation was  normalized; entropy was used to describe texture. Typically, saturation and texture, varied across a pur\' ee flavor's dilution classes and were distinguishable between different pur\' ee flavors.}
\label{fig:sat_entropy}
\end{figure}

\refstepcounter{afigure} 
\begin{figure}
\centering
\includegraphics[width=\textwidth]{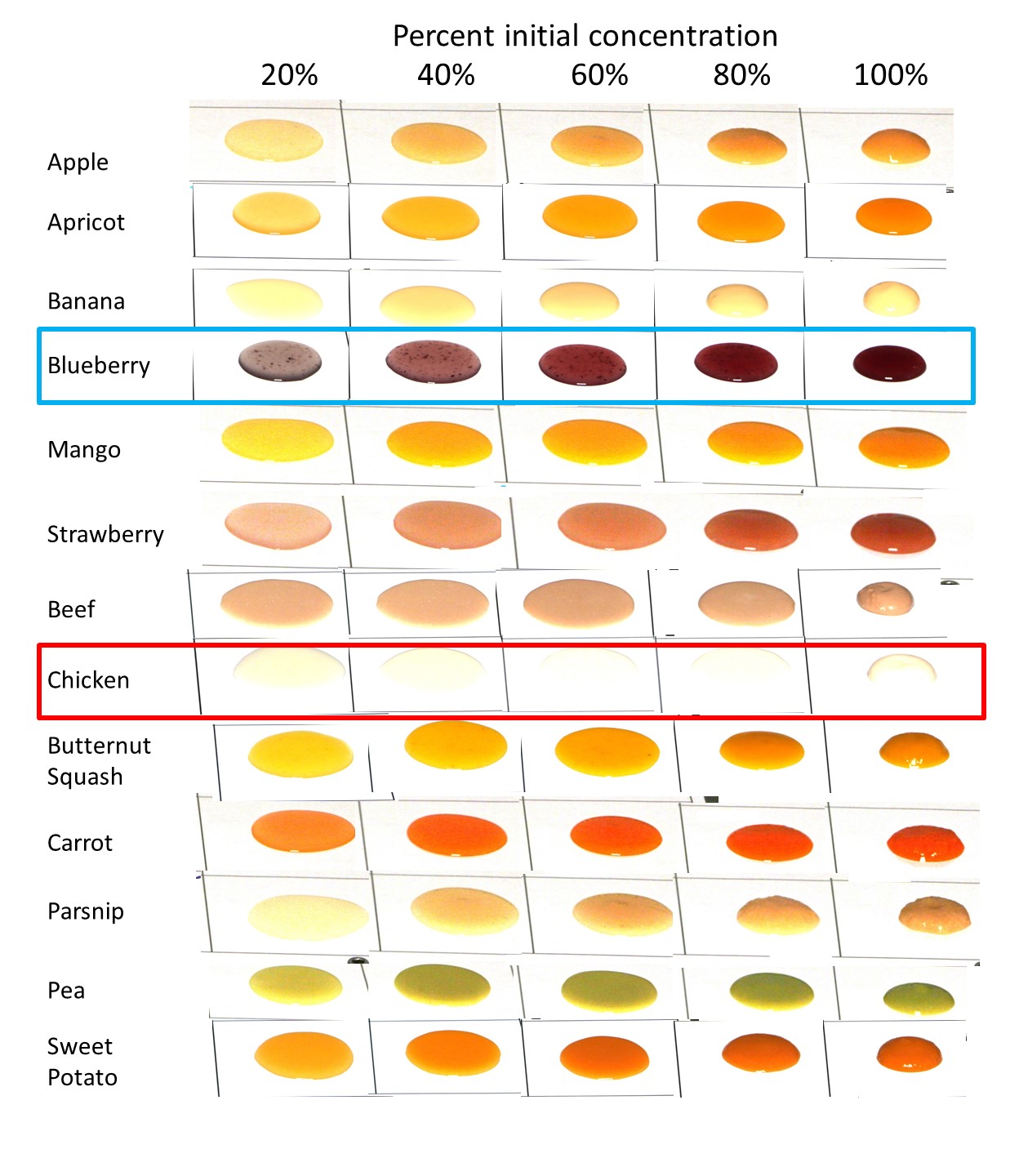}
\caption{Sample horizontal polarized subimages taken with ISO 100, EXP 1/20 s. Note the visible variations across dilutions for the best performing pur\' ee flavor, with the stacked autoencoders and softmax layer, highlighted in blue (average accuracy blueberry: 99.6\%) and the indistinguishable color and lack of texture consistent across the poorest performing pur\' ee flavor highlighted in red (average accuracy chicken: 73.3\%); the subimages at 60\% and 80\% appear nearly absent or completely saturated.}
\label{fig:sample_blobs}
\end{figure}

\refstepcounter{atable} 

\begin{table}[htbp]
  \centering
  \caption{Summary of sensitivity, specificity and accuracy for each flavor using self-generated features extracted from an autoencoder and four methods for to discriminate between dilutions: softmax layer, random forest, SVM - linear kernel SVM, and SVM - radial basis kernel. Results summarized represent five randomly initiated networks with 6-fold cross-validation (i.e., leave one of each of the six imaged positions out for testing) for each flavor.}
\begin{tabular}{lccc}
    \multicolumn{4}{c}{}\\  
    \toprule
          & \multicolumn{3}{c}{\textbf{AUTOENCODER 2 FEATURES}} \\
          & \multicolumn{3}{c}{\textit{\textbf{SVM Linear Kernel}}}\\
    \midrule
    \textbf{Flavour} & \textbf{Sens} & \textbf{Spec} & \textbf{Acc}\\
    & $\mu_{Sens} \pm \sigma_{Sens}$ & $\mu_{Spec} \pm \sigma_{Spec}$ & $\mu_{Acc} \pm \sigma_{Acc}$ \\
    \midrule
    Apple 		 & 0.732 $\pm$ 0.211 & 0.923 $\pm$ 0.065 & 0.880 $\pm$ 0.043 \\
    Apricot 	 & 0.750 $\pm$ 0.112 & 0.927 $\pm$ 0.036 & 0.887 $\pm$ 0.026 \\
    Banana 		 & 0.607 $\pm$ 0.214 & 0.859 $\pm$ 0.119 & 0.792 $\pm$ 0.096 \\
    Beef 		 & 0.608 $\pm$ 0.252 & 0.868 $\pm$ 0.096 & 0.798 $\pm$ 0.070 \\
    Blueberry  	 & 0.840 $\pm$ 0.101 & 0.956 $\pm$ 0.031 & 0.931 $\pm$ 0.030 \\
    Carrot 		 & 0.628 $\pm$ 0.171 & 0.884 $\pm$ 0.068 & 0.820 $\pm$ 0.043 \\
    Chicken  	 & 0.395 $\pm$ 0.322 & 0.781 $\pm$ 0.186 & 0.621 $\pm$ 0.156 \\
    Mango 		 & 0.668 $\pm$ 0.243 & 0.894 $\pm$ 0.080 & 0.841 $\pm$ 0.072 \\
    Parsnip 	 & 0.765 $\pm$ 0.198 & 0.928 $\pm$ 0.071 & 0.889 $\pm$ 0.085 \\
    Pea 		 & 0.657 $\pm$ 0.140 & 0.892 $\pm$ 0.052 & 0.834 $\pm$ 0.037 \\
    Squash 		 & 0.713 $\pm$ 0.195 & 0.911 $\pm$ 0.067 & 0.865 $\pm$ 0.057 \\
    Strawberry 	 & 0.918 $\pm$ 0.063 & 0.979 $\pm$ 0.016 & 0.966 $\pm$ 0.013 \\
    Sweet Potato & 0.688 $\pm$ 0.202 & 0.903 $\pm$ 0.080 & 0.850 $\pm$ 0.062 \\
    \midrule
    \textit{Across Flavours} & \textit{0.690 $\pm$ 0.187} & \textit{0.900 $\pm$ 0.074} & \textit{0.844 $\pm$ 0.061} \\
    \midrule
    \multicolumn{4}{c}{}\\    
    \midrule
    & \multicolumn{3}{c}{\textit{\textbf{SVM Radial Basis Kernel}}}\\
    \midrule
    \textbf{Flavour} & \textbf{Sens} & \textbf{Spec} & \textbf{Acc}\\
    & $\mu_{Sens} \pm \sigma_{Sens}$ & $\mu_{Spec} \pm \sigma_{Spec}$ & $\mu_{Acc} \pm \sigma_{Acc}$ \\
    \midrule
    Apple 		 & 0.635 $\pm$ 0.188 & 0.895 $\pm$ 0.063 & 0.828 $\pm$ 0.061 \\
    Apricot 	 & 0.702 $\pm$ 0.134 & 0.907 $\pm$ 0.052 & 0.859 $\pm$ 0.046 \\
    Banana 		 & 0.509 $\pm$ 0.159 & 0.807 $\pm$ 0.129 & 0.717 $\pm$ 0.105 \\
    Beef 		 & 0.446 $\pm$ 0.267 & 0.795 $\pm$ 0.108 & 0.679 $\pm$ 0.082 \\
    Blueberry 	 & 0.736 $\pm$ 0.096 & 0.923 $\pm$ 0.035 & 0.880 $\pm$ 0.028 \\
    Carrot 		 & 0.573 $\pm$ 0.131 & 0.865 $\pm$ 0.064 & 0.786 $\pm$ 0.049 \\
    Chicken 	 & 0.385 $\pm$ 0.317 & 0.776 $\pm$ 0.190 & 0.609 $\pm$ 0.162 \\
    Mango 		 & 0.655 $\pm$ 0.240 & 0.889 $\pm$ 0.082 & 0.833 $\pm$ 0.078 \\
    Parsnip 	 & 0.627 $\pm$ 0.232 & 0.880 $\pm$ 0.091 & 0.813 $\pm$ 0.107 \\
    Pea 		 & 0.578 $\pm$ 0.128 & 0.851 $\pm$ 0.061 & 0.777 $\pm$ 0.043 \\
    Squash 		 & 0.645 $\pm$ 0.193 & 0.887 $\pm$ 0.081 & 0.827 $\pm$ 0.060 \\
    Strawberry 	 & 0.852 $\pm$ 0.063 & 0.958 $\pm$ 0.016 & 0.935 $\pm$ 0.019 \\
    Sweet Potato & 0.567 $\pm$ 0.181 & 0.854 $\pm$ 0.094 & 0.773 $\pm$ 0.076 \\
    \midrule
    \textit{Across Flavours} & \textit{0.609 $\pm$ 0.179} & \textit{0.868 $\pm$ 0.082} & \textit{0.794 $\pm$ 0.070} \\
    \bottomrule
    \end{tabular}%
  \label{tab:AutoFeat}%
\end{table}%

\begin{table*}[htbp]
  \centering
	\begin{tabular}{lccc}
	\multicolumn{4}{l}{Table~\ref{tab:AutoFeat} continued:}\\
	\multicolumn{4}{c}{}\\   
    \toprule
          & \multicolumn{3}{c}{\textbf{AUTOENCODER 2 FEATURES}} \\
          & \multicolumn{3}{c}{\textit{\textbf{Random Forest}}} \\
    \midrule
    \textbf{Flavour} & \textbf{Sens} & \textbf{Spec} & \textbf{Acc}\\
    & $\mu_{Sens} \pm \sigma_{Sens}$ & $\mu_{Spec} \pm \sigma_{Spec}$ & $\mu_{Acc} \pm \sigma_{Acc}$ \\
    \midrule
    Apple 		 & 0.824 $\pm$ 0.140 & 0.952 $\pm$ 0.039 & 0.925 $\pm$ 0.036 \\
    Apricot 	 & 0.859 $\pm$ 0.097 & 0.962 $\pm$ 0.030 & 0.940 $\pm$ 0.029 \\
    Banana 		 & 0.841 $\pm$ 0.160 & 0.956 $\pm$ 0.043 & 0.931 $\pm$ 0.050 \\
    Beef 		 & 0.708 $\pm$ 0.204 & 0.910 $\pm$ 0.064 & 0.862 $\pm$ 0.046 \\
    Blueberry 	 & 0.906 $\pm$ 0.077 & 0.975 $\pm$ 0.026 & 0.961 $\pm$ 0.027 \\
    Carrot 		 & 0.709 $\pm$ 0.144 & 0.914 $\pm$ 0.062 & 0.866 $\pm$ 0.033 \\
    Chicken 	 & 0.522 $\pm$ 0.206 & 0.826 $\pm$ 0.085 & 0.744 $\pm$ 0.073 \\
    Mango 		 & 0.759 $\pm$ 0.195 & 0.928 $\pm$ 0.060 & 0.890 $\pm$ 0.053 \\
    Parsnip 	 & 0.867 $\pm$ 0.144 & 0.963 $\pm$ 0.039 & 0.942 $\pm$ 0.046 \\
    Pea 		 & 0.728 $\pm$ 0.119 & 0.920 $\pm$ 0.040 & 0.876 $\pm$ 0.032 \\
    Squash 		 & 0.793 $\pm$ 0.175 & 0.941 $\pm$ 0.048 & 0.908 $\pm$ 0.041 \\
    Strawberry 	 & 0.934 $\pm$ 0.066 & 0.983 $\pm$ 0.017 & 0.973 $\pm$ 0.014 \\
    Sweet Potato & 0.813 $\pm$ 0.126 & 0.948 $\pm$ 0.038 & 0.918 $\pm$ 0.032 \\
    \midrule
    \textit{Across Flavours} & \textit{0.790 $\pm$ 0.143} & \textit{0.937 $\pm$ 0.046} & \textit{0.903 $\pm$ 0.039} \\
    \midrule
    \multicolumn{4}{c}{}\\
    \midrule
          & \multicolumn{3}{c}{\textit{\textbf{Softmax Layer}}} \\
    \midrule
    \textbf{Flavour} & \textbf{Sens} & \textbf{Spec} & \textbf{Acc}\\
    & $\mu_{Sens} \pm \sigma_{Sens}$ & $\mu_{Spec} \pm \sigma_{Spec}$ & $\mu_{Acc} \pm \sigma_{Acc}$ \\
    \midrule
    Apple 		 & 0.871 $\pm$ 0.154 & 0.966 $\pm$ 0.041 & 0.947 $\pm$ 0.033 \\
    Apricot 	 & 0.920 $\pm$ 0.073 & 0.979 $\pm$ 0.017 & 0.967 $\pm$ 0.018 \\
    Banana 		 & 0.874 $\pm$ 0.188 & 0.964 $\pm$ 0.054 & 0.944 $\pm$ 0.057 \\
    Beef 		 & 0.709 $\pm$ 0.249 & 0.914 $\pm$ 0.076 & 0.865 $\pm$ 0.063 \\
    Blueberry 	 & 0.989 $\pm$ 0.019 & 0.997 $\pm$ 0.005 & 0.996 $\pm$ 0.006 \\
    Carrot 		 & 0.769 $\pm$ 0.194 & 0.935 $\pm$ 0.074 & 0.897 $\pm$ 0.059 \\
    Chicken 	 & 0.501 $\pm$ 0.286 & 0.824 $\pm$ 0.124 & 0.733 $\pm$ 0.078 \\
    Mango 		 & 0.748 $\pm$ 0.232 & 0.923 $\pm$ 0.074 & 0.883 $\pm$ 0.068 \\
    Parsnip 	 & 0.923 $\pm$ 0.118 & 0.980 $\pm$ 0.033 & 0.968 $\pm$ 0.038 \\
    Pea 		 & 0.765 $\pm$ 0.113 & 0.935 $\pm$ 0.032 & 0.898 $\pm$ 0.024 \\
    Squash 		 & 0.887 $\pm$ 0.138 & 0.969 $\pm$ 0.039 & 0.952 $\pm$ 0.043 \\
    Strawberry 	 & 0.955 $\pm$ 0.048 & 0.988 $\pm$ 0.013 & 0.982 $\pm$ 0.012 \\
    Sweet Potato & 0.884 $\pm$ 0.133 & 0.969 $\pm$ 0.039 & 0.951 $\pm$ 0.035 \\
    \midrule
    \textit{Across Flavours} & \textbf{\textit{0.830 $\pm$ 0.150}} & \textbf{\textit{0.950 $\pm$ 0.048}} & \textbf{\textit{0.922 $\pm$ 0.041}} \\
    \bottomrule
    \end{tabular}%
\end{table*}%

\refstepcounter{atable} 
\begin{table}[htbp]
  \centering
  \caption{Summary of sensitivity, specificity and accuracy for each flavor using 71 handcrafted features pertaining to color and texture and three methods to discriminate between dilutions: random forest, SVM - linear kernel SVM, and SVM - radial basis kernel. Results summarized represent the average from five random forests each containing 10 trees, or one run of SVM for each of linear and radial basis kernels. All methods used 6-fold cross-validation (i.e., leave one of each of the six imaged positions out for testing) for each flavor.}
    \begin{tabular}{lccc}
    \multicolumn{4}{c}{}\\  
    \toprule
    	& \multicolumn{3}{c}{\textbf{HANDCRAFTED FEATURES}} \\
        & \multicolumn{3}{c}{\textit{\textbf{SVM Linear Kernel}}}\\
    \midrule
    \textbf{Flavour} & \textbf{Sens} & \textbf{Spec} & \textbf{Acc}\\
    & $\mu_{Sens} \pm \sigma_{Sens}$ & $\mu_{Spec} \pm \sigma_{Spec}$ & $\mu_{Acc} \pm \sigma_{Acc}$ \\
    \midrule
    Apple 		 & 0.695 $\pm$ 0.150 & 0.908 $\pm$ 0.062 & 0.859 $\pm$ 0.040 \\
    Apricot 	 & 0.731 $\pm$ 0.108 & 0.919 $\pm$ 0.036 & 0.876 $\pm$ 0.015 \\
    Banana 		 & 0.638 $\pm$ 0.270 & 0.876 $\pm$ 0.098 & 0.813 $\pm$ 0.098 \\
    Beef 		 & 0.496 $\pm$ 0.317 & 0.817 $\pm$ 0.178 & 0.720 $\pm$ 0.116 \\
    Blueberry 	 & 0.760 $\pm$ 0.098 & 0.927 $\pm$ 0.030 & 0.888 $\pm$ 0.030 \\
    Carrot 		 & 0.644 $\pm$ 0.183 & 0.889 $\pm$ 0.074 & 0.829 $\pm$ 0.053 \\
    Chicken 	 & 0.398 $\pm$ 0.216 & 0.764 $\pm$ 0.150 & 0.632 $\pm$ 0.068 \\
    Mango 		 & 0.696 $\pm$ 0.172 & 0.907 $\pm$ 0.054 & 0.858 $\pm$ 0.040 \\
    Parsnip 	 & 0.850 $\pm$ 0.151 & 0.958 $\pm$ 0.047 & 0.933 $\pm$ 0.055 \\
    Pea 		 & 0.556 $\pm$ 0.220 & 0.858 $\pm$ 0.111 & 0.774 $\pm$ 0.064 \\
    Squash 		 & 0.741 $\pm$ 0.197 & 0.921 $\pm$ 0.055 & 0.879 $\pm$ 0.070 \\
    Strawberry 	 & 0.864 $\pm$ 0.157 & 0.962 $\pm$ 0.049 & 0.941 $\pm$ 0.036 \\
    Sweet Potato & 0.580 $\pm$ 0.200 & 0.861 $\pm$ 0.112 & 0.787 $\pm$ 0.053 \\
    \midrule
    \textit{Across Flavours} & \textit{0.665 $\pm$ 0.188} & \textit{0.890 $\pm$ 0.081} & \textit{0.830 $\pm$ 0.057} \\
    \midrule
    \multicolumn{4}{c}{}\\
    \midrule
        & \multicolumn{3}{c}{\textit{\textbf{SVM Radial Basis Kernel}}}\\
    \midrule
    \textbf{Flavour} & \textbf{Sens} & \textbf{Spec} & \textbf{Acc}\\
    & $\mu_{Sens} \pm \sigma_{Sens}$ & $\mu_{Spec} \pm \sigma_{Spec}$ & $\mu_{Acc} \pm \sigma_{Acc}$ \\
    \midrule
    Apple 		 & 0.613 $\pm$ 0.225 & 0.870 $\pm$ 0.086 & 0.806 $\pm$ 0.082 \\
    Apricot 	 & 0.729 $\pm$ 0.066 & 0.917 $\pm$ 0.033 & 0.874 $\pm$ 0.025 \\
    Banana 		 & 0.441 $\pm$ 0.219 & 0.780 $\pm$ 0.118 & 0.662 $\pm$ 0.116 \\
    Beef 		 & 0.381 $\pm$ 0.310 & 0.745 $\pm$ 0.151 & 0.612 $\pm$ 0.086 \\
    Blueberry 	 & 0.548 $\pm$ 0.095 & 0.839 $\pm$ 0.059 & 0.756 $\pm$ 0.066 \\
    Carrot 		 & 0.585 $\pm$ 0.118 & 0.865 $\pm$ 0.061 & 0.794 $\pm$ 0.039 \\
    Chicken 	 & 0.332 $\pm$ 0.177 & 0.730 $\pm$ 0.104 & 0.566 $\pm$ 0.066 \\
    Mango 		 & 0.701 $\pm$ 0.125 & 0.911 $\pm$ 0.062 & 0.862 $\pm$ 0.046 \\
    Parsnip 	 & 0.580 $\pm$ 0.180 & 0.859 $\pm$ 0.054 & 0.783 $\pm$ 0.066 \\
    Pea 		 & 0.515 $\pm$ 0.208 & 0.837 $\pm$ 0.111 & 0.746 $\pm$ 0.057 \\
    Squash 		 & 0.731 $\pm$ 0.185 & 0.918 $\pm$ 0.054 & 0.873 $\pm$ 0.070 \\
    Strawberry 	 & 0.806 $\pm$ 0.186 & 0.943 $\pm$ 0.051 & 0.912 $\pm$ 0.049 \\
    Sweet Potato & 0.539 $\pm$ 0.183 & 0.838 $\pm$ 0.075 & 0.758 $\pm$ 0.049 \\
    \midrule
    \textit{Across Flavours} & \textit{0.577 $\pm$ 0.175} & \textit{0.850 $\pm$ 0.078} & \textit{0.770 $\pm$ 0.063} \\
    \bottomrule
    \end{tabular}%
    \label{tab:HandFeat}%
\end{table}%
    
\begin{table*}[htbp]
	\centering
    \begin{tabular}{lccc}
    \multicolumn{4}{l}{Table~\ref{tab:HandFeat} continued:}\\
    \multicolumn{4}{c}{}\\  
    \toprule    
		&  \multicolumn{3}{c}{\textbf{HANDCRAFTED FEATURES}}\\
        &  \multicolumn{3}{c}{\textit{\textbf{Random Forest}}}\\
    \midrule
    \textbf{Flavour} & \textbf{Sens} & \textbf{Spec} & \textbf{Acc} \\
    & $\mu_{Sens} \pm \sigma_{Sens}$ & $\mu_{Spec} \pm \sigma_{Spec}$ & $\mu_{Acc} \pm \sigma_{Acc}$ \\
	\midrule
	Apple 		 & 0.871 $\pm$ 0.170 & 0.966 $\pm$ 0.051 & 0.946 $\pm$ 0.048 \\
    Apricot 	 & 0.923 $\pm$ 0.068 & 0.980 $\pm$ 0.020 & 0.968 $\pm$ 0.017 \\
    Banana 		 & 0.855 $\pm$ 0.177 & 0.960 $\pm$ 0.050 & 0.936 $\pm$ 0.053 \\
    Beef 		 & 0.775 $\pm$ 0.155 & 0.935 $\pm$ 0.043 & 0.900 $\pm$ 0.039 \\
    Blueberry 	 & 0.929 $\pm$ 0.090 & 0.981 $\pm$ 0.022 & 0.971 $\pm$ 0.028 \\
    Carrot 		 & 0.769 $\pm$ 0.173 & 0.935 $\pm$ 0.056 & 0.897 $\pm$ 0.043 \\
    Chicken 	 & 0.532 $\pm$ 0.257 & 0.833 $\pm$ 0.103 & 0.748 $\pm$ 0.087 \\
    Mango 		 & 0.742 $\pm$ 0.241 & 0.923 $\pm$ 0.069 & 0.881 $\pm$ 0.061 \\
    Parsnip 	 & 0.922 $\pm$ 0.141 & 0.979 $\pm$ 0.036 & 0.967 $\pm$ 0.040 \\
    Pea 		 & 0.811 $\pm$ 0.138 & 0.948 $\pm$ 0.047 & 0.917 $\pm$ 0.040 \\
    Squash 		 & 0.882 $\pm$ 0.165 & 0.969 $\pm$ 0.041 & 0.951 $\pm$ 0.047 \\
    Strawberry 	 & 0.908 $\pm$ 0.139 & 0.976 $\pm$ 0.036 & 0.962 $\pm$ 0.035 \\
    Sweet Potato & 0.812 $\pm$ 0.134 & 0.948 $\pm$ 0.043 & 0.919 $\pm$ 0.032 \\
    \midrule
	\textit{Across Flavours}  & \textbf{\textit{0.826 $\pm$ 0.158}} & \textbf{\textit{0.949 $\pm$ 0.047}} & \textbf{\textit{0.920 $\pm$ 0.044}} \\
	\bottomrule
    \end{tabular}%
\end{table*}%

\refstepcounter{atable} 
\begin{sidewaystable}[htbp]
  	\centering
  	\caption{Descriptive statistics of color features: Mean ($\mu \pm \sigma$) red (R), green (G), blue (B), and saturation (S) values for each flavor across 20\% (most diluted), 40\%, 60\%, 80\%, and 100\% (undiluted) concentrations.}
\begin{tabular}{llccccc}
	\multicolumn{7}{c}{}\\ 
	\toprule
    \multicolumn{1}{r}{} & & \multicolumn{5}{c}{\textbf{PERCENT INITIAL CONCENTRATION}} \\
    \midrule
    \multicolumn{2}{l}{\textbf{Flavour (FRUIT)}} & \textbf{20\%} & \textbf{40\%} & \textbf{60\%} & \textbf{80\%} & \textbf{100\%} \\
    \midrule
    Apple 		 & R ($\mu_R \pm \sigma_R$) & 0.981 $\pm$ 0.0228 & 0.976 $\pm$ 0.0279 & 0.966 $\pm$ 0.0385 & 0.957 $\pm$ 0.0578 & 0.919 $\pm$ 0.0816 \\
       			 & G ($\mu_G \pm \sigma_G$) & 0.879 $\pm$ 0.0497 & 0.719 $\pm$ 0.0482 & 0.633 $\pm$ 0.0518 & 0.587 $\pm$ 0.0726 & 0.523 $\pm$ 0.0678 \\
         		 & B ($\mu_B \pm \sigma_B$) & 0.489 $\pm$ 0.0581 & 0.261 $\pm$ 0.0577 & 0.192 $\pm$ 0.054 & 0.164 $\pm$ 0.0521 & 0.147 $\pm$ 0.0471 \\
         		 & S ($\mu_S \pm \sigma_S$) & 0.865 $\pm$ 0.0358 & 0.743 $\pm$ 0.035 & 0.683 $\pm$ 0.0393 & 0.649 $\pm$ 0.0568 & 0.598 $\pm$ 0.0614 \\
	\midrule
	Apricot 	 & R ($\mu_R \pm \sigma_R$) & 0.993 $\pm$ 0.00945 & 0.994 $\pm$ 0.00754 & 0.994 $\pm$ 0.00958 & 0.991 $\pm$ 0.015 & 0.98 $\pm$ 0.0355 \\
    			 & G ($\mu_G \pm \sigma_G$) & 0.851 $\pm$ 0.0483 & 0.733 $\pm$ 0.0508 & 0.631 $\pm$ 0.0581 & 0.546 $\pm$ 0.0618 & 0.469 $\pm$ 0.0632 \\
          		 & B ($\mu_B \pm \sigma_B$) & 0.377 $\pm$ 0.0643 & 0.154 $\pm$ 0.057 & 0.0656 $\pm$ 0.0455 & 0.0402 $\pm$ 0.0346 & 0.0269 $\pm$ 0.0315 \\
          		 & S ($\mu_S \pm \sigma_S$) & 0.839 $\pm$ 0.0347 & 0.745 $\pm$ 0.0351 & 0.675 $\pm$ 0.038 & 0.622 $\pm$ 0.0396 & 0.571 $\pm$ 0.0428 \\
    \midrule
	Banana		 & R ($\mu_R \pm \sigma_R$) & 0.997 $\pm$ 0.00461 & 0.994 $\pm$ 0.00705 & 0.987 $\pm$ 0.0247 & 0.991 $\pm$ 0.0231 & 0.994 $\pm$ 0.0155 \\
         		 & G ($\mu_G \pm \sigma_G$) & 0.983 $\pm$ 0.023 & 0.897 $\pm$ 0.0475 & 0.855 $\pm$ 0.0748 & 0.901 $\pm$ 0.0775 & 0.939 $\pm$ 0.0759 \\
          		 & B ($\mu_B \pm \sigma_B$) & 0.631 $\pm$ 0.058 & 0.507 $\pm$ 0.0585 & 0.499 $\pm$ 0.0631 & 0.518 $\pm$ 0.0662 & 0.526 $\pm$ 0.0722 \\
          		 & S ($\mu_S \pm \sigma_S$) & 0.947 $\pm$ 0.0187 & 0.882 $\pm$ 0.033 & 0.854 $\pm$ 0.0534 & 0.884 $\pm$ 0.0549 & 0.908 $\pm$ 0.0528 \\
    \midrule
	Blueberry	 & R ($\mu_R \pm \sigma_R$) & 0.622 $\pm$ 0.0796 & 0.592 $\pm$ 0.0694 & 0.498 $\pm$ 0.068 & 0.392 $\pm$ 0.0642 & 0.273 $\pm$ 0.0583 \\
         		 & G ($\mu_G \pm \sigma_G$) & 0.512 $\pm$ 0.0686 & 0.367 $\pm$ 0.0572 & 0.196 $\pm$ 0.0481 & 0.111 $\pm$ 0.0422 & 0.0668 $\pm$ 0.0352 \\
          		 & B ($\mu_B \pm \sigma_B$) & 0.449 $\pm$ 0.0779 & 0.328 $\pm$ 0.0665 & 0.179 $\pm$ 0.0527 & 0.11 $\pm$ 0.0461 & 0.0656 $\pm$ 0.0362 \\
 				 & S ($\mu_S \pm \sigma_S$) & 0.538 $\pm$ 0.0709 & 0.43 $\pm$ 0.0592 & 0.284 $\pm$ 0.0505 & 0.195 $\pm$ 0.0431 & 0.128 $\pm$ 0.0345 \\
	\midrule	
	Mango		 & R ($\mu_R \pm \sigma_R$) & 0.99 $\pm$ 0.0103 & 0.991 $\pm$ 0.00995 & 0.99 $\pm$ 0.0111 & 0.984 $\pm$ 0.0215 & 0.96 $\pm$ 0.0492 \\
         		 & G ($\mu_G \pm \sigma_G$) & 0.844 $\pm$ 0.0491 & 0.647 $\pm$ 0.0455 & 0.606 $\pm$ 0.0447 & 0.551 $\pm$ 0.0517 & 0.486 $\pm$ 0.0475 \\
          		 & B ($\mu_B \pm \sigma_B$) & 0.153 $\pm$ 0.0587 & 0.0429 $\pm$ 0.0381 & 0.0383 $\pm$ 0.0328 & 0.0337 $\pm$ 0.0317 & 0.038 $\pm$ 0.0355 \\
          		 & S ($\mu_S \pm \sigma_S$) & 0.809 $\pm$ 0.0317 & 0.681 $\pm$ 0.028 & 0.656 $\pm$ 0.0277 & 0.621 $\pm$ 0.0327 & 0.577 $\pm$ 0.0365 \\
    \midrule
	Strawberry	 & R ($\mu_R \pm \sigma_R$) & 0.933 $\pm$ 0.0466 & 0.879 $\pm$ 0.0554 & 0.855 $\pm$ 0.0565 & 0.73 $\pm$ 0.0663 & 0.634 $\pm$ 0.0775 \\
         		 & G ($\mu_G \pm \sigma_G$) & 0.726 $\pm$ 0.0454 & 0.552 $\pm$ 0.0446 & 0.474 $\pm$ 0.0444 & 0.291 $\pm$ 0.0425 & 0.23 $\pm$ 0.0396 \\
          		 & B ($\mu_B \pm \sigma_B$) & 0.542 $\pm$ 0.0551 & 0.348 $\pm$ 0.0519 & 0.269 $\pm$ 0.0496 & 0.127 $\pm$ 0.0458 & 0.102 $\pm$ 0.0412 \\
          		 & S ($\mu_S \pm \sigma_S$) & 0.767 $\pm$ 0.0415 & 0.627 $\pm$ 0.0399 & 0.565 $\pm$ 0.0394 & 0.403 $\pm$ 0.0417 & 0.336 $\pm$ 0.0412 \\    
    \bottomrule
    \end{tabular}%
  \label{tab:RGB_sat_ent}%
\end{sidewaystable}%

\begin{sidewaystable*}[htbp]
  	\centering
	\begin{tabular}{llccccc}
	\multicolumn{4}{l}{Table~\ref{tab:RGB_sat_ent} continued:}\\ 
	\multicolumn{4}{c}{}\\ 
	\toprule
    \multicolumn{1}{r}{} & & \multicolumn{5}{c}{\textbf{PERCENT INITIAL CONCENTRATION}} \\
	\multicolumn{2}{l}{\textbf{Flavour (MEAT)}} & \textbf{20\%} & \textbf{40\%} & \textbf{60\%} & \textbf{80\%} & \textbf{100\%} \\	
	\midrule
	    Beef		 & R ($\mu_R \pm \sigma_R$) & 0.966 $\pm$ 0.0345 & 0.907 $\pm$ 0.0564 & 0.849 $\pm$ 0.0607 & 0.855 $\pm$ 0.0613 & 0.928 $\pm$ 0.0842 \\
         		 & G ($\mu_G \pm \sigma_G$) & 0.715 $\pm$ 0.0482 & 0.621 $\pm$ 0.0451 & 0.603 $\pm$ 0.0456 & 0.622 $\pm$ 0.0451 & 0.672 $\pm$ 0.0832 \\
          		 & B ($\mu_B \pm \sigma_B$) & 0.52 $\pm$ 0.0503 & 0.432 $\pm$ 0.0484 & 0.425 $\pm$ 0.0499 & 0.444 $\pm$ 0.0486 & 0.466 $\pm$ 0.0684 \\
          		 & S ($\mu_S \pm \sigma_S$) & 0.768 $\pm$ 0.0398 & 0.685 $\pm$ 0.0433 & 0.656 $\pm$ 0.0456 & 0.672 $\pm$ 0.045 & 0.725 $\pm$ 0.0766 \\
    \midrule
    Chicken		 & R ($\mu_R \pm \sigma_R$) & 0.998 $\pm$ 0.00329 & 0.997 $\pm$ 0.00401 & 0.997 $\pm$ 0.0045 & 0.996 $\pm$ 0.00506 & 0.997 $\pm$ 0.00417 \\
         		 & G ($\mu_G \pm \sigma_G$) & 0.999 $\pm$ 0.00273 & 0.998 $\pm$ 0.00431 & 0.995 $\pm$ 0.00904 & 0.993 $\pm$ 0.0134 & 0.997 $\pm$ 0.0123 \\
          		 & B ($\mu_B \pm \sigma_B$) & 0.925 $\pm$ 0.0646 & 0.863 $\pm$ 0.0602 & 0.826 $\pm$ 0.0575 & 0.815 $\pm$ 0.0579 & 0.918 $\pm$ 0.0844 \\
          		 & S ($\mu_S \pm \sigma_S$) & 0.99 $\pm$ 0.00826 & 0.982 $\pm$ 0.00813 & 0.977 $\pm$ 0.0106 & 0.973 $\pm$ 0.0132 & 0.988 $\pm$ 0.0153 \\    
    \midrule
    \multicolumn{2}{l}{\textbf{Flavour (VEGETABLE)}} & \textbf{20\%} & \textbf{40\%} & \textbf{60\%} & \textbf{80\%} & \textbf{100\%} \\
    \midrule
    Carrot		 & R ($\mu_R \pm \sigma_R$) & 0.991 $\pm$ 0.00947 & 0.983 $\pm$ 0.0196 & 0.966 $\pm$ 0.0405 & 0.958 $\pm$ 0.0549 & 0.911 $\pm$ 0.0863 \\
         		 & G ($\mu_G \pm \sigma_G$) & 0.523 $\pm$ 0.0473 & 0.33 $\pm$ 0.0452 & 0.284 $\pm$ 0.0411 & 0.25 $\pm$ 0.0373 & 0.245 $\pm$ 0.0353 \\
          		 & B ($\mu_B \pm \sigma_B$) & 0.14 $\pm$ 0.0576 & 0.0485 $\pm$ 0.0386 & 0.0455 $\pm$ 0.0415 & 0.0362 $\pm$ 0.0341 & 0.0389 $\pm$ 0.0352 \\
          		 & S ($\mu_S \pm \sigma_S$) & 0.619 $\pm$ 0.0326 & 0.493 $\pm$ 0.0296 & 0.461 $\pm$ 0.0292 & 0.437 $\pm$ 0.0289 & 0.421 $\pm$ 0.0362 \\
    \midrule
    Parsnip		 & R ($\mu_R \pm \sigma_R$) & 0.99 $\pm$ 0.0144 & 0.967 $\pm$ 0.0329 & 0.929 $\pm$ 0.0631 & 0.967 $\pm$ 0.0487 & 0.869 $\pm$ 0.108 \\
         		 & G ($\mu_G \pm \sigma_G$) & 0.952 $\pm$ 0.0425 & 0.727 $\pm$ 0.042 & 0.638 $\pm$ 0.0541 & 0.761 $\pm$ 0.127 & 0.576 $\pm$ 0.0908 \\
          		 & B ($\mu_B \pm \sigma_B$) & 0.621 $\pm$ 0.0636 & 0.341 $\pm$ 0.0562 & 0.276 $\pm$ 0.0545 & 0.429 $\pm$ 0.248 & 0.264 $\pm$ 0.0782 \\
          		 & S ($\mu_S \pm \sigma_S$) & 0.926 $\pm$ 0.0316 & 0.755 $\pm$ 0.0325 & 0.684 $\pm$ 0.0487 & 0.785 $\pm$ 0.107 & 0.628 $\pm$ 0.0871 \\
    \midrule
    Pea			 & R ($\mu_R \pm \sigma_R$) & 0.832 $\pm$ 0.0514 & 0.65 $\pm$ 0.0541 & 0.624 $\pm$ 0.0533 & 0.553 $\pm$ 0.0587 & 0.58 $\pm$ 0.072 \\
         		 & G ($\mu_G \pm \sigma_G$) & 0.761 $\pm$ 0.0372 & 0.594 $\pm$ 0.0364 & 0.577 $\pm$ 0.0371 & 0.541 $\pm$ 0.0409 & 0.569 $\pm$ 0.0547 \\
          		 & B ($\mu_B \pm \sigma_B$) & 0.236 $\pm$ 0.058 & 0.182 $\pm$ 0.0509 & 0.187 $\pm$ 0.0513 & 0.181 $\pm$ 0.0495 & 0.176 $\pm$ 0.0468 \\
          		 & S ($\mu_S \pm \sigma_S$) & 0.722 $\pm$ 0.0336 & 0.564 $\pm$ 0.0347 & 0.547 $\pm$ 0.0355 & 0.504 $\pm$ 0.0401 & 0.528 $\pm$ 0.0523 \\
    \midrule         
    Squash		 & R ($\mu_R \pm \sigma_R$) & 0.992 $\pm$ 0.00784 & 0.993 $\pm$ 0.00709 & 0.992 $\pm$ 0.01 & 0.95 $\pm$ 0.0629 & 0.943 $\pm$ 0.0796 \\
         		 & G ($\mu_G \pm \sigma_G$) & 0.787 $\pm$ 0.0521 & 0.661 $\pm$ 0.0474 & 0.565 $\pm$ 0.0484 & 0.481 $\pm$ 0.0614 & 0.491 $\pm$ 0.0783 \\
          		 & B ($\mu_B \pm \sigma_B$) & 0.111 $\pm$ 0.0481 & 0.0342 $\pm$ 0.0282 & 0.0289 $\pm$ 0.0241 & 0.0291 $\pm$ 0.028 & 0.032 $\pm$ 0.0508 \\
          		 & S ($\mu_S \pm \sigma_S$) & 0.771 $\pm$ 0.0326 & 0.689 $\pm$ 0.0292 & 0.631 $\pm$ 0.03 & 0.57 $\pm$ 0.0499 & 0.574 $\pm$ 0.0644 \\
    \midrule       
    Sweet Potato & R ($\mu_R \pm \sigma_R$) & 0.992 $\pm$ 0.00826 & 0.984 $\pm$ 0.02 & 0.911 $\pm$ 0.0643 & 0.904 $\pm$ 0.0851 & 0.917 $\pm$ 0.0866 \\
         		 & G ($\mu_G \pm \sigma_G$) & 0.59 $\pm$ 0.0514 & 0.435 $\pm$ 0.0481 & 0.362 $\pm$ 0.0427 & 0.359 $\pm$ 0.0447 & 0.358 $\pm$ 0.0448 \\
          		 & B ($\mu_B \pm \sigma_B$) & 0.0966 $\pm$ 0.0451 & 0.0563 $\pm$ 0.0363 & 0.0658 $\pm$ 0.0388 & 0.0498 $\pm$ 0.0305 & 0.0439 $\pm$ 0.0304 \\
          		 & S ($\mu_S \pm \sigma_S$) & 0.654 $\pm$ 0.0337 & 0.556 $\pm$ 0.0329 & 0.493 $\pm$ 0.039 & 0.487 $\pm$ 0.0456 & 0.489 $\pm$ 0.046 \\
    \bottomrule
    \end{tabular}%
\end{sidewaystable*}%

\refstepcounter{atable} 
\begin{sidewaystable}[htbp]
  	\centering
  	\caption{Descriptive statistics of computational texture features: Mean ($\mu$), Contrast ($c_n$), Homogeneity ($h_g$), Energy ($e_n$), Variance ($\sigma^2$), Correlation ($c_r$), and Entropy ($h_n$) values for each flavor across 20\% (most diluted), 40\%, 60\%, 80\%, and 100\% (undiluted) concentrations. See equations (\ref{eq:mean})--(\ref{eq:entropy}). All values are expressed as ($\mu \pm \sigma$). For display purposes, $\text{Contrast}=c_n\times 10^4$ and $\text{Energy}=e_n\times 10^4$.}
\begin{tabular}{llccccc}
	\multicolumn{7}{c}{}\\ 
	\toprule
    \multicolumn{1}{r}{} & & \multicolumn{5}{c}{\textbf{PERCENT INITIAL CONCENTRATION}} \\
    \midrule
    \multicolumn{2}{l}{\textbf{Flavour (FRUIT)}} & \textbf{20\%} & \textbf{40\%} & \textbf{60\%} & \textbf{80\%} & \textbf{100\%} \\
    \midrule
Apple & Mean & 1.73 $\pm$ 0.0609 & 1.49 $\pm$ 0.0553 & 1.37 $\pm$ 0.0482 & 1.3 $\pm$ 0.0597 & 1.2 $\pm$ 0.0671 \\ 
& Contrast & 2.48 $\pm$ 2.45 & 2.66 $\pm$ 2.61 & 2.54 $\pm$ 2.53 & 2.5 $\pm$ 2.44 & 2.9 $\pm$ 2.94 \\ 
& Homogeneity & 1 $\pm$ 0.000244 & 1 $\pm$ 0.00026 & 1 $\pm$ 0.000253 & 1 $\pm$ 0.000243 & 1 $\pm$ 0.000293 \\ 
& Energy & 9.75 $\pm$ 2.9 & 9.1 $\pm$ 1.67 & 9.04 $\pm$ 3.14 & 7.98 $\pm$ 1.65 & 7.48 $\pm$ 1.4 \\ 
& Variance & 1.5 $\pm$ 0.105 & 1.11 $\pm$ 0.0825 & 0.936 $\pm$ 0.0661 & 0.851 $\pm$ 0.0771 & 0.724 $\pm$ 0.08 \\ 
& Correlation & 1.5 $\pm$ 0.105 & 1.11 $\pm$ 0.0825 & 0.935 $\pm$ 0.0662 & 0.851 $\pm$ 0.0771 & 0.724 $\pm$ 0.08 \\ 
& Entropy & 3.17 $\pm$ 0.0797 & 3.19 $\pm$ 0.0577 & 3.2 $\pm$ 0.0823 & 3.23 $\pm$ 0.0704 & 3.25 $\pm$ 0.0625 \\ 
    \midrule   
Apricot & Mean & 1.68 $\pm$ 0.0648 & 1.49 $\pm$ 0.0666 & 1.35 $\pm$ 0.0735 & 1.24 $\pm$ 0.0739 & 1.14 $\pm$ 0.074 \\ 
& Contrast & 0.856 $\pm$ 0.674 & 0.834 $\pm$ 0.593 & 0.637 $\pm$ 0.495 & 0.654 $\pm$ 0.506 & 0.764 $\pm$ 0.626 \\ 
& Homogeneity & 1 $\pm$ 6.73e-05 & 1 $\pm$ 5.92e-05 & 1 $\pm$ 4.95e-05 & 1 $\pm$ 5.06e-05 & 1 $\pm$ 6.24e-05 \\ 
& Energy & 10.9 $\pm$ 3.99 & 12.5 $\pm$ 5.64 & 15.2 $\pm$ 8.47 & 13.4 $\pm$ 8.99 & 14.4 $\pm$ 15.7 \\ 
& Variance & 1.41 $\pm$ 0.109 & 1.11 $\pm$ 0.101 & 0.916 $\pm$ 0.101 & 0.777 $\pm$ 0.0938 & 0.657 $\pm$ 0.0866 \\ 
& Correlation & 1.41 $\pm$ 0.109 & 1.11 $\pm$ 0.101 & 0.916 $\pm$ 0.101 & 0.777 $\pm$ 0.0938 & 0.657 $\pm$ 0.0866 \\ 
& Entropy & 3.16 $\pm$ 0.0809 & 3.13 $\pm$ 0.077 & 3.07 $\pm$ 0.127 & 3.12 $\pm$ 0.135 & 3.11 $\pm$ 0.204 \\ 
    \midrule   
Banana & Mean & 1.9 $\pm$ 0.0331 & 1.76 $\pm$ 0.0568 & 1.71 $\pm$ 0.0709 & 1.77 $\pm$ 0.0611 & 1.82 $\pm$ 0.0433 \\ 
& Contrast & 0.252 $\pm$ 0.381 & 0.775 $\pm$ 0.632 & 0.959 $\pm$ 0.708 & 0.885 $\pm$ 0.691 & 0.875 $\pm$ 0.75 \\ 
& Homogeneity & 1 $\pm$ 3.81e-05 & 1 $\pm$ 6.31e-05 & 1 $\pm$ 7.07e-05 & 1 $\pm$ 6.91e-05 & 1 $\pm$ 7.48e-05 \\ 
& Energy & 85.8 $\pm$ 212 & 13.7 $\pm$ 9.19 & 8.94 $\pm$ 2.73 & 12.4 $\pm$ 5.13 & 53.8 $\pm$ 43.9 \\ 
& Variance & 1.8 $\pm$ 0.0623 & 1.56 $\pm$ 0.0998 & 1.46 $\pm$ 0.12 & 1.57 $\pm$ 0.105 & 1.66 $\pm$ 0.076 \\ 
& Correlation & 1.8 $\pm$ 0.0623 & 1.56 $\pm$ 0.0999 & 1.46 $\pm$ 0.12 & 1.57 $\pm$ 0.105 & 1.66 $\pm$ 0.0761 \\ 
& Entropy & 2.83 $\pm$ 0.432 & 3.13 $\pm$ 0.103 & 3.23 $\pm$ 0.0692 & 3.15 $\pm$ 0.0962 & 2.82 $\pm$ 0.207 \\ 
 \midrule   
Blueberry & Mean & 1.07 $\pm$ 0.115 & 0.86 $\pm$ 0.1 & 0.568 $\pm$ 0.093 & 0.39 $\pm$ 0.081 & 0.256 $\pm$ 0.0599 \\ 
& Contrast & 3.59 $\pm$ 2.51 & 2.97 $\pm$ 1.74 & 1.7 $\pm$ 0.899 & 1.17 $\pm$ 1.11 & 0.884 $\pm$ 0.568 \\ 
& Homogeneity & 1 $\pm$ 0.000249 & 1 $\pm$ 0.000173 & 1 $\pm$ 8.91e-05 & 1 $\pm$ 0.000103 & 1 $\pm$ 5.67e-05 \\ 
& Energy & 23.2 $\pm$ 17.1 & 16.9 $\pm$ 7.87 & 21.6 $\pm$ 28.7 & 17.7 $\pm$ 34.9 & 12.5 $\pm$ 5.55 \\ 
& Variance & 0.588 $\pm$ 0.127 & 0.377 $\pm$ 0.0912 & 0.166 $\pm$ 0.0547 & 0.0796 $\pm$ 0.032 & 0.0352 $\pm$ 0.0159 \\ 
& Correlation & 0.588 $\pm$ 0.127 & 0.376 $\pm$ 0.0912 & 0.166 $\pm$ 0.0548 & 0.0795 $\pm$ 0.032 & 0.0351 $\pm$ 0.0159 \\ 
& Entropy & 2.94 $\pm$ 0.218 & 3 $\pm$ 0.14 & 2.98 $\pm$ 0.249 & 3.07 $\pm$ 0.229 & 3.11 $\pm$ 0.105 \\ 
          \bottomrule
    \end{tabular}%
  \label{tab:appendix_texture_features}%
\end{sidewaystable}%

\begin{sidewaystable*}[htbp]
  	\centering
	\begin{tabular}{llccccc}
	\multicolumn{4}{l}{Table~\ref{tab:appendix_texture_features} continued:}\\ 
	\multicolumn{4}{c}{}\\ 
	\toprule
    \multicolumn{1}{r}{} & & \multicolumn{5}{c}{\textbf{PERCENT INITIAL CONCENTRATION}} \\
		\multicolumn{2}{l}{\textbf{Flavour (FRUIT CONTINUED)}} & \textbf{20\%} & \textbf{40\%} & \textbf{60\%} & \textbf{80\%} & \textbf{100\%} \\	
 \midrule 
Mango & Mean & 1.62 $\pm$ 0.0526 & 1.36 $\pm$ 0.0452 & 1.31 $\pm$ 0.0428 & 1.24 $\pm$ 0.0451 & 1.15 $\pm$ 0.0432 \\ 
& Contrast & 2.41 $\pm$ 2.43 & 1.75 $\pm$ 1.76 & 1.6 $\pm$ 1.54 & 1.91 $\pm$ 1.88 & 2.35 $\pm$ 2.45 \\ 
& Homogeneity & 1 $\pm$ 0.000242 & 1 $\pm$ 0.000176 & 1 $\pm$ 0.000153 & 1 $\pm$ 0.000187 & 1 $\pm$ 0.000244 \\ 
& Energy & 9.19 $\pm$ 1.45 & 10.6 $\pm$ 2.27 & 10.4 $\pm$ 2.26 & 9.02 $\pm$ 1.62 & 8.37 $\pm$ 1.39 \\ 
& Variance & 1.31 $\pm$ 0.0849 & 0.93 $\pm$ 0.0618 & 0.864 $\pm$ 0.0566 & 0.775 $\pm$ 0.0564 & 0.669 $\pm$ 0.0501 \\ 
& Correlation & 1.31 $\pm$ 0.085 & 0.93 $\pm$ 0.0618 & 0.863 $\pm$ 0.0567 & 0.775 $\pm$ 0.0564 & 0.669 $\pm$ 0.0501 \\ 
& Entropy & 3.18 $\pm$ 0.0531 & 3.15 $\pm$ 0.0613 & 3.15 $\pm$ 0.0623 & 3.19 $\pm$ 0.0586 & 3.22 $\pm$ 0.0567 \\ 
	\midrule 
Strawberry & Mean & 1.53 $\pm$ 0.0737 & 1.25 $\pm$ 0.0674 & 1.13 $\pm$ 0.0644 & 0.806 $\pm$ 0.0528 & 0.671 $\pm$ 0.0459 \\ 
& Contrast & 3.79 $\pm$ 3.88 & 3.81 $\pm$ 3.88 & 3.72 $\pm$ 3.96 & 3.28 $\pm$ 3.54 & 3.17 $\pm$ 3.42 \\ 
& Homogeneity & 1 $\pm$ 0.000387 & 1 $\pm$ 0.000386 & 1 $\pm$ 0.000395 & 1 $\pm$ 0.000353 & 1 $\pm$ 0.000341 \\ 
& Energy & 9.23 $\pm$ 1.33 & 9.16 $\pm$ 1.53 & 9.02 $\pm$ 1.55 & 8.19 $\pm$ 1.23 & 8.4 $\pm$ 1.58 \\ 
& Variance & 1.18 $\pm$ 0.113 & 0.788 $\pm$ 0.0846 & 0.641 $\pm$ 0.0727 & 0.328 $\pm$ 0.0424 & 0.229 $\pm$ 0.0313 \\ 
& Correlation & 1.18 $\pm$ 0.113 & 0.788 $\pm$ 0.0847 & 0.64 $\pm$ 0.0728 & 0.328 $\pm$ 0.0425 & 0.228 $\pm$ 0.0314 \\ 
& Entropy & 3.18 $\pm$ 0.0523 & 3.18 $\pm$ 0.0578 & 3.19 $\pm$ 0.0601 & 3.23 $\pm$ 0.0532 & 3.23 $\pm$ 0.0578 \\ 
	    \midrule 
	\multicolumn{2}{l}{\textbf{Flavour (MEAT)}} & \textbf{20\%} & \textbf{40\%} & \textbf{60\%} & \textbf{80\%} & \textbf{100\%} \\	
	\midrule   
Beef & Mean & 1.54 $\pm$ 0.073 & 1.37 $\pm$ 0.0794 & 1.31 $\pm$ 0.083 & 1.34 $\pm$ 0.077 & 1.45 $\pm$ 0.0773 \\ 
& Contrast & 2.34 $\pm$ 1.7 & 2.58 $\pm$ 2.1 & 2.58 $\pm$ 2.13 & 2.56 $\pm$ 2.1 & 3.03 $\pm$ 1.87 \\ 
& Homogeneity & 1 $\pm$ 0.00017 & 1 $\pm$ 0.00021 & 1 $\pm$ 0.000213 & 1 $\pm$ 0.000209 & 1 $\pm$ 0.000187 \\ 
& Energy & 16.5 $\pm$ 11.1 & 11.9 $\pm$ 7.34 & 9.6 $\pm$ 2.62 & 9.28 $\pm$ 2 & 10.7 $\pm$ 4.53 \\ 
& Variance & 1.18 $\pm$ 0.112 & 0.942 $\pm$ 0.109 & 0.865 $\pm$ 0.11 & 0.906 $\pm$ 0.104 & 1.06 $\pm$ 0.112 \\ 
& Correlation & 1.18 $\pm$ 0.112 & 0.942 $\pm$ 0.109 & 0.864 $\pm$ 0.11 & 0.906 $\pm$ 0.104 & 1.06 $\pm$ 0.112 \\ 
& Entropy & 3 $\pm$ 0.182 & 3.12 $\pm$ 0.154 & 3.17 $\pm$ 0.0877 & 3.19 $\pm$ 0.0696 & 3.16 $\pm$ 0.131 \\ 
 
   \midrule  
Chicken & Mean & 1.98 $\pm$ 0.0145 & 1.97 $\pm$ 0.0132 & 1.95 $\pm$ 0.0178 & 1.95 $\pm$ 0.0227 & 1.98 $\pm$ 0.0163 \\ 
& Contrast & 0.0721 $\pm$ 0.101 & 0.125 $\pm$ 0.149 & 0.209 $\pm$ 0.329 & 0.305 $\pm$ 0.521 & 0.178 $\pm$ 0.255 \\ 
& Homogeneity & 1 $\pm$ 1.01e-05 & 1 $\pm$ 1.49e-05 & 1 $\pm$ 3.29e-05 & 1 $\pm$ 5.21e-05 & 1 $\pm$ 2.55e-05 \\ 
& Energy & 587 $\pm$ 1.59e+03 & 31.1 $\pm$ 47.5 & 46.1 $\pm$ 95.5 & 30.5 $\pm$ 52.3 & 308 $\pm$ 337 \\ 
& Variance & 1.96 $\pm$ 0.0288 & 1.93 $\pm$ 0.0259 & 1.91 $\pm$ 0.0347 & 1.9 $\pm$ 0.0439 & 1.95 $\pm$ 0.0317 \\ 
& Correlation & 1.96 $\pm$ 0.0288 & 1.93 $\pm$ 0.0259 & 1.91 $\pm$ 0.0347 & 1.9 $\pm$ 0.0439 & 1.95 $\pm$ 0.0318 \\ 
& Entropy & 2.48 $\pm$ 0.761 & 2.92 $\pm$ 0.238 & 2.89 $\pm$ 0.313 & 2.94 $\pm$ 0.252 & 2.47 $\pm$ 0.372 \\ 
   \bottomrule
    \end{tabular}%
  \label{tab:texture_features}%
\end{sidewaystable*}%
 
\begin{sidewaystable*}[htbp]
  	\centering
	\begin{tabular}{llccccc}
	\multicolumn{4}{l}{Table~\ref{tab:appendix_texture_features} continued:}\\ 
	\multicolumn{4}{c}{}\\ 
	\toprule
    \multicolumn{1}{r}{} & & \multicolumn{5}{c}{\textbf{PERCENT INITIAL CONCENTRATION}} \\
 \multicolumn{2}{l}{\textbf{Flavour (VEGETABLES)}} & \textbf{20\%} & \textbf{40\%} & \textbf{60\%} & \textbf{80\%} & \textbf{100\%} \\	
	\midrule    
Carrot & Mean & 1.24 $\pm$ 0.0575 & 0.986 $\pm$ 0.0511 & 0.922 $\pm$ 0.0467 & 0.876 $\pm$ 0.0407 & 0.842 $\pm$ 0.0411 \\ 
& Contrast & 1.9 $\pm$ 2.08 & 2.33 $\pm$ 2.39 & 2.35 $\pm$ 2.42 & 2.19 $\pm$ 2.22 & 2.77 $\pm$ 2.62 \\ 
& Homogeneity & 1 $\pm$ 0.000208 & 1 $\pm$ 0.000239 & 1 $\pm$ 0.000241 & 1 $\pm$ 0.000222 & 1 $\pm$ 0.000261 \\ 
& Energy & 10.1 $\pm$ 2.36 & 10.8 $\pm$ 2.43 & 10 $\pm$ 1.95 & 10.2 $\pm$ 2.1 & 10.3 $\pm$ 7.56 \\ 
& Variance & 0.771 $\pm$ 0.0715 & 0.488 $\pm$ 0.0503 & 0.427 $\pm$ 0.043 & 0.385 $\pm$ 0.0356 & 0.357 $\pm$ 0.0344 \\ 
& Correlation & 0.771 $\pm$ 0.0715 & 0.488 $\pm$ 0.0503 & 0.426 $\pm$ 0.0431 & 0.385 $\pm$ 0.0356 & 0.357 $\pm$ 0.0345 \\ 
& Entropy & 3.15 $\pm$ 0.0663 & 3.14 $\pm$ 0.068 & 3.16 $\pm$ 0.0605 & 3.15 $\pm$ 0.0695 & 3.17 $\pm$ 0.142 \\ 
    \midrule   
Parsnip & Mean & 1.85 $\pm$ 0.0552 & 1.51 $\pm$ 0.0507 & 1.37 $\pm$ 0.0589 & 1.57 $\pm$ 0.163 & 1.26 $\pm$ 0.091 \\ 
& Contrast & 1.63 $\pm$ 2.01 & 3.12 $\pm$ 3.01 & 3.35 $\pm$ 3.17 & 3.2 $\pm$ 2.55 & 5.58 $\pm$ 4.46 \\ 
& Homogeneity & 1 $\pm$ 0.000201 & 1 $\pm$ 0.0003 & 1 $\pm$ 0.000315 & 1 $\pm$ 0.000254 & 0.999 $\pm$ 0.000443 \\ 
& Energy & 14.8 $\pm$ 12.9 & 9.69 $\pm$ 2.11 & 8.06 $\pm$ 1.27 & 343 $\pm$ 1.25e+03 & 16.1 $\pm$ 14.5 \\ 
& Variance & 1.72 $\pm$ 0.102 & 1.14 $\pm$ 0.0765 & 0.939 $\pm$ 0.0799 & 1.25 $\pm$ 0.276 & 0.804 $\pm$ 0.115 \\ 
& Correlation & 1.72 $\pm$ 0.102 & 1.14 $\pm$ 0.0767 & 0.939 $\pm$ 0.08 & 1.25 $\pm$ 0.276 & 0.804 $\pm$ 0.115 \\ 
& Entropy & 3.08 $\pm$ 0.173 & 3.17 $\pm$ 0.0605 & 3.24 $\pm$ 0.0535 & 3 $\pm$ 0.628 & 3.07 $\pm$ 0.244 \\ 
    \midrule   
Pea & Mean & 1.44 $\pm$ 0.0458 & 1.13 $\pm$ 0.058 & 1.09 $\pm$ 0.0608 & 1.01 $\pm$ 0.0652 & 1.05 $\pm$ 0.0696 \\ 
& Contrast & 3.85 $\pm$ 3.88 & 3.75 $\pm$ 3.91 & 3.72 $\pm$ 3.98 & 3.64 $\pm$ 4.03 & 3.74 $\pm$ 3.99 \\ 
& Homogeneity & 1 $\pm$ 0.000386 & 1 $\pm$ 0.000389 & 1 $\pm$ 0.000396 & 1 $\pm$ 0.000401 & 1 $\pm$ 0.000397 \\ 
& Energy & 8.96 $\pm$ 1.43 & 9.94 $\pm$ 1.65 & 9.84 $\pm$ 1.55 & 9.5 $\pm$ 1.53 & 8.52 $\pm$ 1.7 \\ 
& Variance & 1.04 $\pm$ 0.0661 & 0.639 $\pm$ 0.0657 & 0.6 $\pm$ 0.0666 & 0.51 $\pm$ 0.0658 & 0.562 $\pm$ 0.0742 \\ 
& Correlation & 1.04 $\pm$ 0.0663 & 0.638 $\pm$ 0.0659 & 0.6 $\pm$ 0.0668 & 0.51 $\pm$ 0.0659 & 0.561 $\pm$ 0.0744 \\ 
& Entropy & 3.19 $\pm$ 0.0566 & 3.15 $\pm$ 0.0572 & 3.15 $\pm$ 0.0576 & 3.17 $\pm$ 0.0546 & 3.22 $\pm$ 0.0612 \\ 
    \midrule   
Squash & Mean & 1.54 $\pm$ 0.0576 & 1.38 $\pm$ 0.0525 & 1.26 $\pm$ 0.0546 & 1.14 $\pm$ 0.0674 & 1.15 $\pm$ 0.0665 \\ 
& Contrast & 1.52 $\pm$ 1.38 & 1.09 $\pm$ 0.874 & 1.11 $\pm$ 1.08 & 1.44 $\pm$ 1.28 & 3.47 $\pm$ 3.07 \\ 
& Homogeneity & 1 $\pm$ 0.000138 & 1 $\pm$ 8.73e-05 & 1 $\pm$ 0.000108 & 1 $\pm$ 0.000127 & 1 $\pm$ 0.000301 \\ 
& Energy & 10.6 $\pm$ 3.59 & 13 $\pm$ 7.97 & 12.1 $\pm$ 4.56 & 8.51 $\pm$ 1.65 & 31.6 $\pm$ 31.9 \\ 
& Variance & 1.19 $\pm$ 0.0889 & 0.952 $\pm$ 0.0733 & 0.8 $\pm$ 0.0701 & 0.655 $\pm$ 0.0768 & 0.667 $\pm$ 0.0771 \\ 
& Correlation & 1.19 $\pm$ 0.089 & 0.951 $\pm$ 0.0734 & 0.8 $\pm$ 0.0702 & 0.655 $\pm$ 0.0768 & 0.667 $\pm$ 0.0771 \\ 
& Entropy & 3.15 $\pm$ 0.0849 & 3.1 $\pm$ 0.124 & 3.11 $\pm$ 0.0945 & 3.22 $\pm$ 0.0591 & 2.91 $\pm$ 0.37 \\ 
    \midrule   
SweetPotato & Mean & 1.31 $\pm$ 0.0623 & 1.11 $\pm$ 0.0616 & 0.985 $\pm$ 0.0677 & 0.974 $\pm$ 0.0683 & 0.979 $\pm$ 0.0617 \\ 
& Contrast & 1.2 $\pm$ 1.17 & 1.4 $\pm$ 1.4 & 1.8 $\pm$ 1.63 & 1.81 $\pm$ 1.56 & 1.81 $\pm$ 1.43 \\ 
& Homogeneity & 1 $\pm$ 0.000117 & 1 $\pm$ 0.00014 & 1 $\pm$ 0.000162 & 1 $\pm$ 0.000156 & 1 $\pm$ 0.000143 \\ 
& Energy & 10.6 $\pm$ 2.81 & 12 $\pm$ 12.6 & 9.31 $\pm$ 1.77 & 8.24 $\pm$ 1.33 & 9.79 $\pm$ 6.87 \\ 
& Variance & 0.859 $\pm$ 0.0827 & 0.621 $\pm$ 0.0697 & 0.488 $\pm$ 0.0676 & 0.478 $\pm$ 0.0671 & 0.483 $\pm$ 0.0611 \\ 
& Correlation & 0.859 $\pm$ 0.0827 & 0.621 $\pm$ 0.0698 & 0.488 $\pm$ 0.0677 & 0.478 $\pm$ 0.0672 & 0.483 $\pm$ 0.0612 \\ 
& Entropy & 3.15 $\pm$ 0.0649 & 3.12 $\pm$ 0.131 & 3.19 $\pm$ 0.0607 & 3.23 $\pm$ 0.0556 & 3.19 $\pm$ 0.141 \\ 

    \bottomrule
    \end{tabular}%
  \label{tab:texture_features}%
\end{sidewaystable*}%

\end{document}